\documentclass[Afour,sageh,times]{sagej}

\usepackage{moreverb,url}

\usepackage[colorlinks,bookmarksopen,bookmarksnumbered,citecolor=red,urlcolor=red]{hyperref}

\usepackage[latin1]{inputenc}
\usepackage{amsmath}
\usepackage{amsfonts}
\usepackage{amssymb}
\usepackage{graphicx}
\usepackage{pgfplots}
\usepackage{bm}
\usepackage{hhline}
\usepackage{array}
\pgfplotsset{compat=1.10}
\usetikzlibrary{arrows,arrows.meta}
\usepackage{natbib} 
\usepackage{hyperref}
\usepackage[british]{babel}
\usepackage{enumitem}
\usepackage[at]{easylist}
\usepackage{lipsum}
\usepackage{xargs}
\usepackage{xcolor}
\usepackage[colorinlistoftodos,prependcaption,textsize=footnotesize]{todonotes}
\usepackage{flushend}
\usepackage{tabularx}
\usepackage{tikz}
\usepackage{siunitx}
\usepackage{booktabs}

\def \state {s}
\def \action {u}
\def \obser {{o}}

\def \setObser {\mathcal{O}}

\newcommandx{\TODO}[2][1=]{\todo[linecolor=red,backgroundcolor=red!25,bordercolor=red,#1]{#2}}
\newcommandx{\info}[2][1=]{\todo[linecolor=blue,backgroundcolor=blue!25,bordercolor=blue,#1]{#2}}
\newcommandx{\discussion}[2][1=]{\todo[linecolor=OliveGreen,backgroundcolor=OliveGreen!25,bordercolor=OliveGreen,#1]{#2}}
\newcommandx{\unsure}[2][1=]{\todo[linecolor=Plum,backgroundcolor=Plum!25,bordercolor=Plum,#1]{#2}}
\newcommandx{\thiswillnotshow}[2][1=]{\todo[disable,#1]{#2}}
\newcommand\BibTeX{{\rmfamily B\kern-.05em \textsc{i\kern-.025em b}\kern-.08em
T\kern-.1667em\lower.7ex\hbox{E}\kern-.125emX}}

\def\volumeyear{2019}
\usepackage{etoolbox}
\makeatletter
\patchcmd\@combinedblfloats{\box\@outputbox}{\unvbox\@outputbox}{}{%
  \errmessage{\noexpand\@combinedblfloats could not be patched}%
}%
\makeatother 

\begin{document}

\runninghead{Rudenko et al.}

\title{Human Motion Trajectory Prediction: A Survey}

\author{Andrey Rudenko\affilnum{1,2}, Luigi Palmieri\affilnum{1}, Michael Herman\affilnum{3}, Kris M. Kitani\affilnum{4}, Dariu M. Gavrila\affilnum{5} and Kai O. Arras\affilnum{1}}

\affiliation{\affilnum{1}Robert Bosch GmbH, Corporate Research, Germany\\
\affilnum{2}Mobile Robotics and Olfaction Lab, \"Orebro University, Sweden\\
\affilnum{3}Bosch Center for Artificial Intelligence, Germany\\
\affilnum{4}Carnegie Mellon University, USA\\
\affilnum{5}Intelligent Vehicles group, TU Delft, The Netherlands}

\corrauth{Andrey Rudenko, Bosch Corporate Research, Renningen, Germany.}

\email{andrey.rudenko@de.bosch.com}

\begin{abstract}
With growing numbers of intelligent autonomous systems in human environments, the ability of such systems to perceive, understand and anticipate human behavior becomes increasingly important. Specifically, predicting future positions of dynamic agents and planning considering such predictions are key tasks for self-driving vehicles, service robots and advanced surveillance systems.

This paper provides a survey of human motion trajectory prediction. We review, analyze and structure a large selection of work from different communities and propose a taxonomy that categorizes existing methods based on the motion modeling approach and level of contextual information used. We provide an overview of the existing datasets and performance metrics. We discuss limitations of the state of the art and outline directions for further research.
\end{abstract}


\maketitle

\section{Introduction}
\label{sec:introduction}

\begin{figure}[t]
	\includegraphics[height=0.334\linewidth,keepaspectratio]{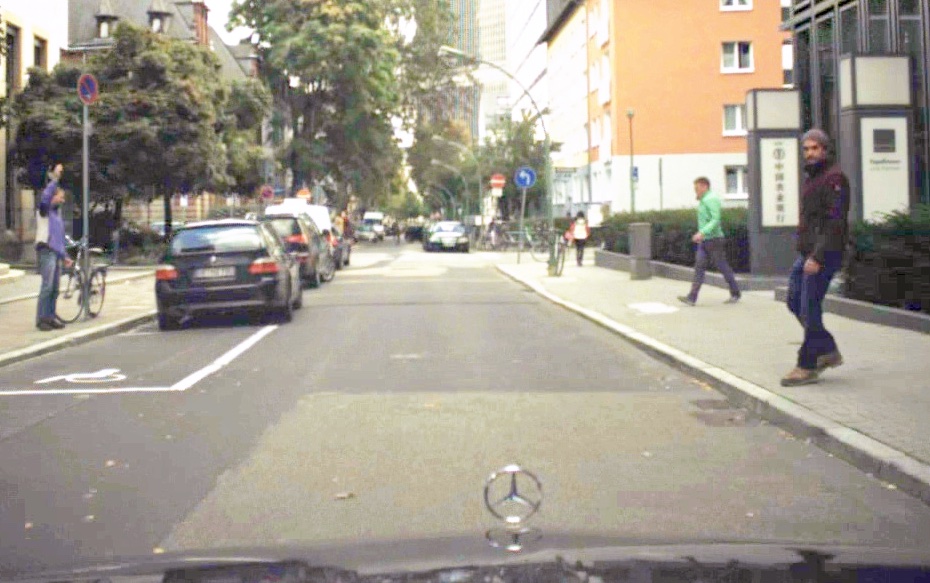} \hfill
	\includegraphics[height=0.334\linewidth,keepaspectratio]{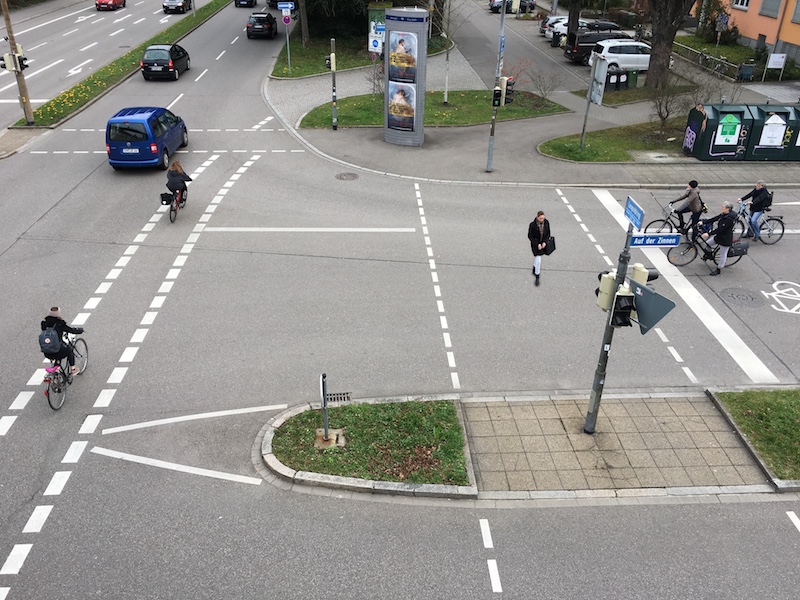}
	
	\vspace{2mm}
	
    \includegraphics[height=0.299\linewidth,keepaspectratio]{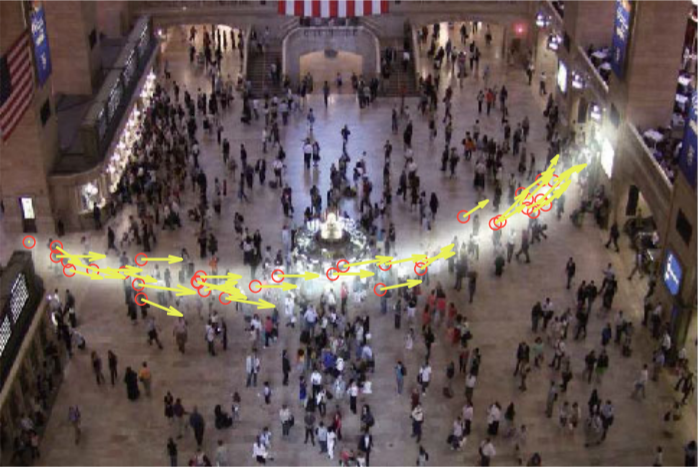}
    \hfill
	\includegraphics[height=0.299\linewidth,keepaspectratio]{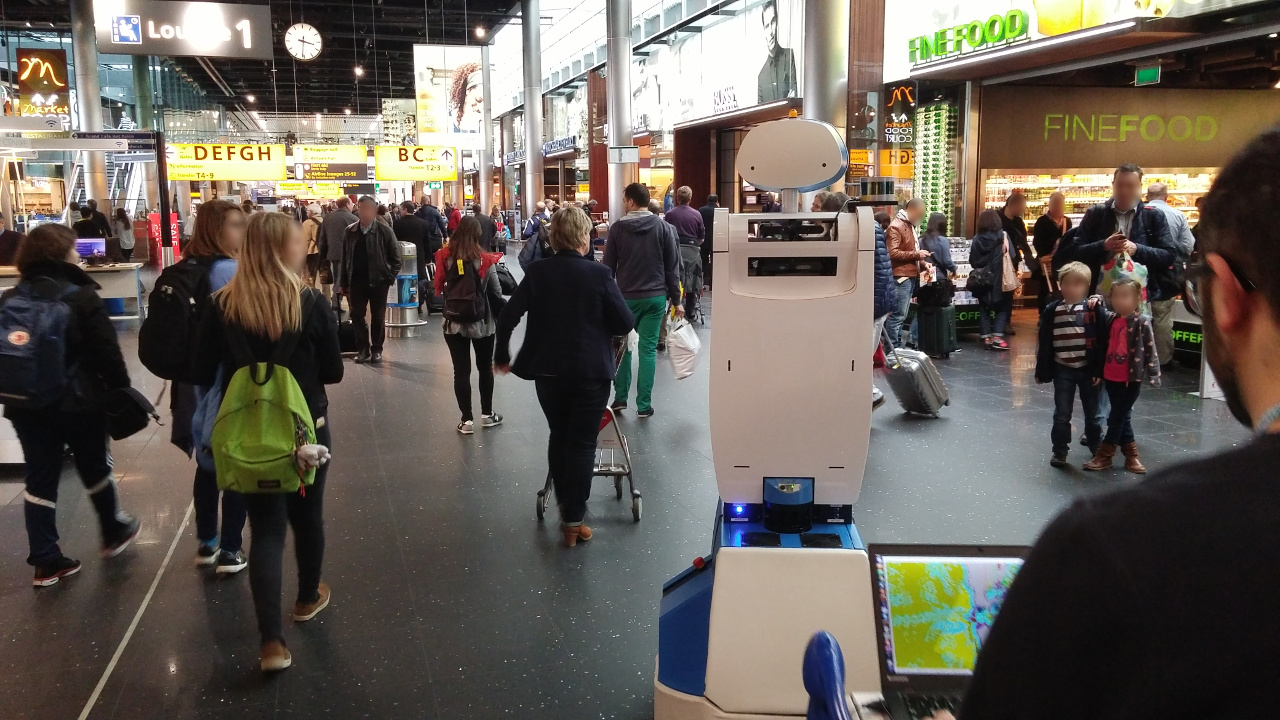}
	\caption{Application domains of human motion prediction. {\bf Top left:} Will the pedestrian cross? Self-driving vehicles have to quickly reason about intentions and future locations of other traffic participants, such as pedestrians (Illustration from \citep{kooij2018ijcv}). {\bf Top right:} Advanced traffic surveillance systems can provide real-time alerts of pending collisions using communication technology. {\bf Bottom left:} Advanced surveillance systems analyze human motion in public spaces for suspicious activity detection or crowd control (Illustration from \citep{zhou2015learning}). {\bf Bottom right:} Robot navigation in densely populated spaces requires accurate motion prediction of surrounding people to safely and efficiently move through crowds.}
	\label{introduction:fig:application-domains}
\end{figure}

Understanding human motion is a key skill for intelligent systems to coexist and interact with humans. It involves aspects in representation, perception and motion analysis. Prediction plays an important part in human motion analysis: foreseeing how a scene involving multiple agents will unfold over time allows to incorporate this knowledge in a pro-active manner, i.e. allowing for enhanced ways of active perception, predictive planning, model predictive control, or human-robot interaction. As such, human motion prediction has received increased attention in recent years across several communities. Many important application domains exist, such as self-driving vehicles, service robots, and advanced surveillance systems, see Fig.~\ref{introduction:fig:application-domains}.


%

The challenge of making accurate predictions of human motion arises from the complexity of human behavior and the variety of its internal and external stimuli. Motion behavior may be driven by own goal intent, the presence and actions of surrounding agents, social relations between agents, social rules and norms, or the environment with its topology, geometry, affordances and semantics. Most factors are not directly observable and need to be inferred from noisy perceptual cues or modeled from context information. 
Furthermore, to be effective in practice, motion prediction should be robust and operate in real-time.

Human motion comes in many forms: articulated full body motion, gestures and facial expressions, or movement through space by walking, using a mobility device or driving a vehicle. The scope of this survey is human motion trajectory prediction. Specifically, we focus on ground-level 2D trajectory prediction for pedestrians and also consider the literature on cyclists and vehicles. Prediction of video frames, articulated motion, or human actions or activities is out of scope although many of those tasks rely on the same motion modeling principles and trajectory prediction methods considered here. Within this scope, we survey a large selection of works from different communities and propose a novel taxonomy based on the motion modeling approaches and the contextual cues. We categorize the state of the art and discuss typical properties, advantages and drawbacks of the categories as well as outline open challenges for future research. Finally, we raise three questions: \emph{Q1}: are the evaluation techniques to measure prediction performance good enough and follow best practices? \emph{Q2}: have all prediction methods arrived on the same performance level and the choice of the modeling approach does not matter anymore? \emph{Q3}: is motion prediction solved?

The paper is structured as follows: we present the taxonomy in Sec.~\ref{sec:taxonomy}, review and analyze the literature on human motion prediction first by the modeling approaches in Sec.~\ref{sec:posterior_distribution:physics_based} -- Sec.~\ref{sec:posterior_distribution:planning_based}, and then by the contextual cues in Sec.~\ref{sec:classification_contextualcues}. In Sec.~\ref{sec:evaluation} we review the benchmarking of motion prediction techniques in terms of commonly used performance metrics and datasets. In Sec.~\ref{sec:discussion} we discuss the state of the art with respect to the above three questions and outline open research challenges. Finally, Sec.~\ref{sec:conclusions} concludes the paper.

We recommend Sec.~\ref{sec:introduction},\ref{sec:taxonomy}, Fig.~\ref{fig:pictograms-physics-based}--\ref{fig:pictograms-planning-based} and Sec.~\ref{sec:discussion} as a coarse overview of the motion prediction methodology for a general reader. A practitioner may find value in the review of the datasets and metrics in Sec.~\ref{sec:evaluation}. Finally, the thorough analysis of the literature in Sec.~\ref{sec:posterior_distribution:physics_based}--\ref{sec:classification_contextualcues} is recommended for expert readers.

\subsection{Overview and Terminology}
\begin{figure}[t]
	\centering
	\includegraphics[width=0.99\columnwidth]{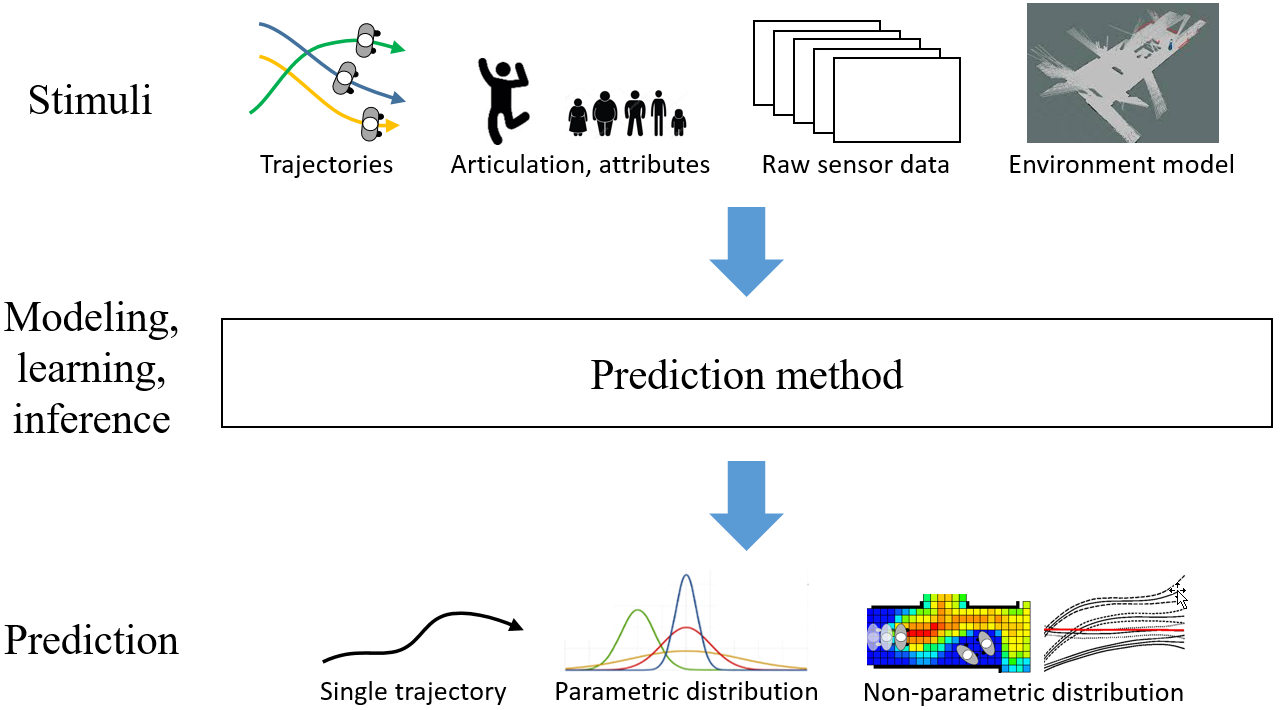}
	\caption{Typical elements of a motion prediction system: internal and external stimuli that influence motion behavior, the method itself and the different parametric, non-parametric or structured forms of predictions.}
	\label{introduction:fig:overview}
\end{figure}

On the highest level of abstraction, the motion prediction problem contains the following three elements (Fig. \ref{introduction:fig:overview}): 
\begin{itemize}
	\item \emph{Stimuli:} Internal and external stimuli that determine motion behavior include the agents' motion intent and other directly or indirectly observable influences. Most prediction methods rely on observed partial trajectories, or generally, sequences of agent state observations such as positions, velocities, body joint angles or attributes. Often, this is provided by a target tracking system and it is common to assume correct track identity over the observation period. Other forms of inputs include contextual cues from the environment such as scene geometry, semantics, or cues that relate to other moving entities in the surrounding. End-to-end approaches rely on sequences of raw sensor data.
	
	\item \emph{Modeling approach:} Approaches to human motion prediction differ in the way they represent, parametrize, learn and solve the task. This paper focuses on finding and analyzing useful categories, hidden similarities, common assumptions and best evaluation practices in the growing body of literature.
	
	\item \emph{Prediction:} Different methods produce different parametric, non-parametric or structured forms of predictions such as Gaussians over agent states, probability distributions over grids, singular or multiple trajectory samples or motion patterns using graphical models. 
\end{itemize}

We use the term {\em agent} to denote dynamic objects of interest such as robots, pedestrians, cyclists, cars or other human-driven vehicles. The {\em target agent} is the dynamic object for which we make the actual motion prediction. We assume the agent behavior to be non-erratic and goal-directed with regard to an optimal or near-optimal expected outcome. This assumption is typical as the motion prediction problem were much harder or even ill-posed otherwise. 
We define a \emph{path} to be a sequence of $(x,y)$-positions and a \emph{trajectory} to be a path combined with a timing law or a velocity profile.
We refer to {\em short-term} and {\em long-term} prediction to characterize prediction horizons of 1-2 $s$ and up to 20 $s$ ahead, respectively.

Formally, we denote $\mathbf{\state}_t$ as the state of an agent at time $t$, $\mathbf{\action}_t$ as the action that the agent takes at time $t$, $\mathbf{\obser}_t \in \setObser$ as the observations of the agent's state at time $t$, and use $\zeta$ to denote trajectories. We refer to a history of several states, actions or observations from time $t$ to time $T$ using subscripts ${t:T}$.

\subsection{Application Domains}
Motion prediction is a key task for service robots, self-driving vehicles, and advanced surveillance systems (Fig.~\ref{introduction:fig:application-domains}). 
\subsubsection{Service robots}
\label{sec:introduction:mobile-robotics}
Mobile service robots increasingly operate in open-ended domestic, industrial and urban environments shared with humans. Anticipating motion of surrounding agents is an important prerequisite for safe and efficient motion planning and human-robot interaction. Limited on-board resources for computation and first-person sensing makes this a challenging task.  

\subsubsection{Self-driving vehicles}
\label{sec:introduction:self-driving-vehicles}
The ability to anticipate motion of other road users is essential for automated driving. Similar challenges apply as in the service robot domain, although they are more pronounced given the higher masses and velocities of vehicles and the resulting larger harm that can potentially be inflicted, especially towards vulnerable road users (i.e. pedestrians and cyclists). Furthermore, vehicles need to operate in rapidly changing, semantically rich outdoor traffic settings and need hard real-time operating constraints. Knowledge of the traffic infrastructure (location of lanes, curbside, traffic signs, traffic lights, other road markings such as zebras) and the traffic rules can help in the motion prediction.

\subsubsection{Surveillance}
\label{sec:discussion:computer-vision}
Visual surveillance of vehicular traffic or human crowds relies on the ability to accurately track a large number of targets across distributed networks of stationary cameras. Long-term motion prediction can support a variety of surveillance tasks such as person retrieval, perimeter protection, traffic monitoring, crowd management or retail analytics by further reducing the number of false positive tracks and track identifier switches, particularly in dense crowds or across non-overlapping fields of views. 

\begin{figure*}[t]
	\centering
	\includegraphics[width=0.99\linewidth]{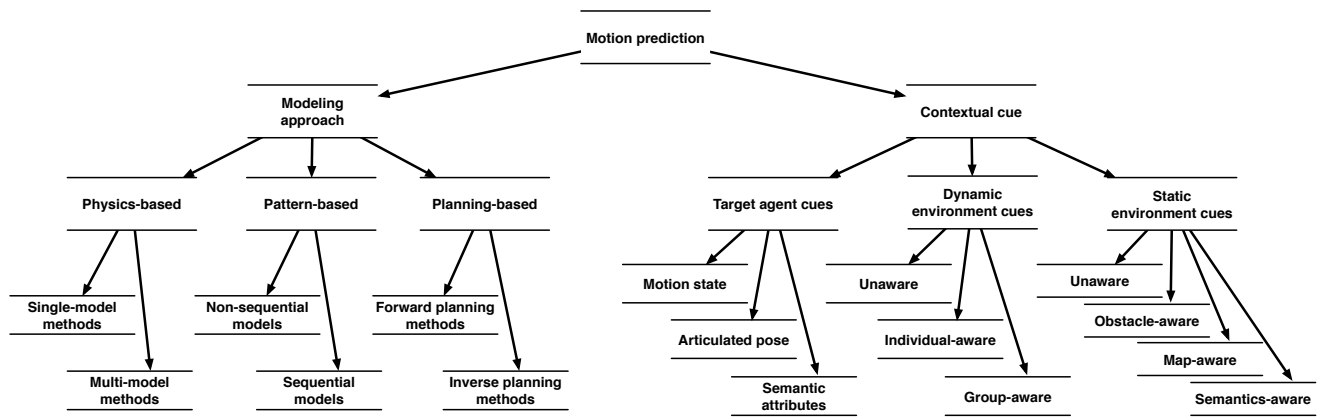}
	\caption{Overview of the categories in our taxonomy.}
	\label{fig:taxonomy-overview}
\end{figure*}

\begin{figure}[t]
	\centering
	\includegraphics[width=0.999\linewidth,keepaspectratio]{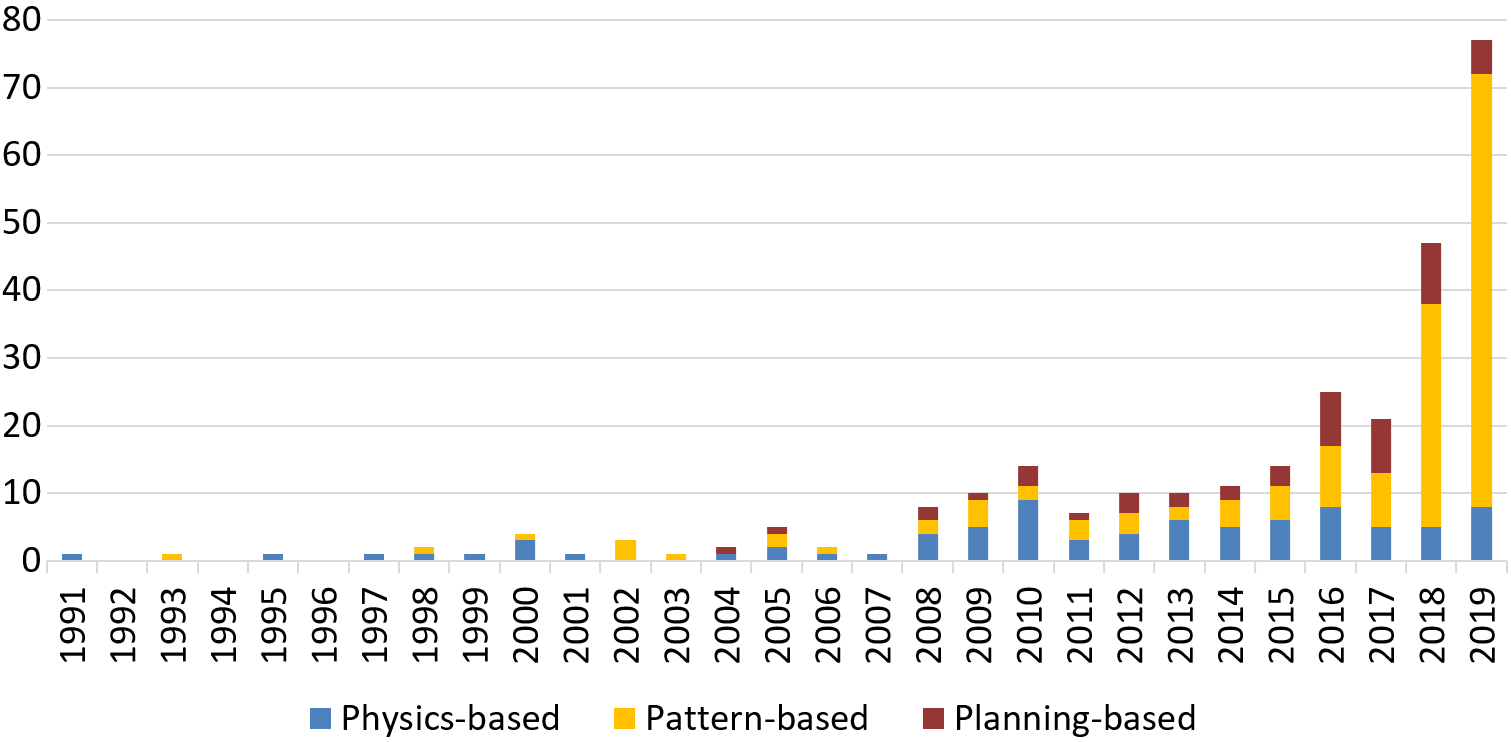}
	\caption{Publications trends in the literature reviewed for this survey, color-coded by modeling approach.
	}
	\label{fig:paperstatistics}
\end{figure}

\subsection{Related Surveys}
%
In this section, we detail related surveys from different scientific communities, i.e. 
robotics \citep{kruse2013human,chik2016review,lasota2017survey}, intelligent vehicles \citep{lefevre2014survey,brouwer2016comparison,ridel2018literature}, and computer vision \citep{morris2008survey,murino2017crowdBehavior,hirakawaDAPI18}.

\cite{kruse2013human} provide a survey of approaches for wheeled mobile robots and categorize human-aware motion based on comfort, naturalness and sociability features. Motion prediction is seen as part of a human-aware navigation framework and categorized into {\em reasoning-based} and {\em learning-based} approaches. In reasoning-based methods, predictions are 
based on simple geometric reasoning or dynamic models of the target agent. 
Learning-based approaches make predictions via motion patterns that are learned from observed agent trajectories.

A short survey on frameworks for socially-aware robot navigation is provided by \cite{chik2016review}. The authors discuss key components of such frameworks including several planners and human motion prediction techniques.

\cite{lasota2017survey} survey the literature on safe human-robot interaction along the four themes of safety through control, motion planning, prediction and psychological factors. In addition to wheeled robots, they also include related works on manipulator arms, drones or self-driving vehicles. The literature on human motion prediction is divided into methods based on {\em goal intent} or {\em motion characteristics}. Goal intent techniques infer an agent's goal and predict a trajectory that the agent is likely to take to reach that goal. The latter group of approaches does not rely explicitly on goals and makes use of observations about how humans move and plan natural paths.

\cite{lefevre2014survey} survey vehicular motion prediction and risk assessment in an automated driving context. The authors discuss the literature based on the semantics used to define motion and risk and distinguish {\em physics-based}, {\em maneuver-based} and {\em interaction-aware}  models for prediction. Physics-based methods predict future trajectories via forward simulation of a vehicle model, typically under kinodynamic constraints and uncertainties in initial states and controls. Maneuver-based methods assume that vehicle motion is a series of typical motion patterns (maneuvers) that have been acquired a priori and can be recognized from observed partial agent trajectories.
Intention-aware methods make joint predictions that account for inter-vehicle interactions, also considering that such interactions are regulated by traffic rules.

\cite{brouwer2016comparison} review and compare pedestrian motion models for vehicle safety systems. According to the cues from the environment used as input for motion prediction,
authors distinguish four classes of methods: \emph{dynamics-based models} which only use the target agent's motion state, methods which use \emph{psychological knowledge of human behavior} in urban environments (e.g. probabilities of acceleration, deceleration, switch of the dynamical model), methods which use \emph{head orientation} and \emph{semantic map} of the environment. This categorization is extended by \cite{ridel2018literature} to review pedestrian crossing intention inference techniques. 

\cite{morris2008survey} survey methods for trajectory learning and analysis for visual surveillance. They discuss similarity metrics, techniques and models for learning prototypical motion patterns (called activity paths) and briefly consider trajectory prediction as a case of online activity analysis. \cite{murino2017crowdBehavior} discuss group and crowd motion analysis as a multidisciplinary problem that combines insights from the social sciences with concepts from computer vision and pattern recognition. The authors review several recent methods for tracking and prediction of human motion in crowds. \cite{hirakawaDAPI18} survey  video-based methods for semantic feature extraction and human trajectory prediction. The literature is divided based on the motion modeling approach into
\emph{Bayesian models}, \emph{energy minimization methods}, \emph{deep learning methods}, \emph{inverse reinforcement learning methods} and \emph{other} approaches.



{Related to our discussion of the benchmarking practices, several works survey the datasets of motion trajectories \citep{poiesi2015predicting,hirakawaDAPI18,ridel2018literature} and metrics for prediction evaluation \citep{quehl2017howgood}. \cite{poiesi2015predicting} and \cite{hirakawaDAPI18} describe several datasets of human trajectories in crowded scenarios, used to study social interactions and evaluate path prediction algorithms. \cite{ridel2018literature} discuss available datasets of pedestrian motion in urban settings. \cite{quehl2017howgood} review several trajectory similarity metrics, applicable in the motion prediction context.
}

Unlike these surveys, we review and analyze the literature across multiple application domains and agent types. Our taxonomy offers a novel way to structure the growing body of literature, containing the categories proposed by \cite{kruse2013human}, \cite{lasota2017survey} and \cite{lefevre2014survey} and extending them with a systematic categorization of contextual cues. In particular, we argue that the modeling approach and the contextual cues are two fundamentally different aspects underlying the motion prediction problem and should be considered separate dimensions for the categorization of methods. This allows, for example, the distinction of physics-based methods that are unaware of any external stimuli from  methods in the same category that are highly situational aware accounting for road geometry, semantics and the presence of other agents. This is unlike previous surveys whose categorizations are along a single dimension based on both different modeling approaches and increasing levels of contextual awareness.

We {extend existing reviews of the}
benchmarking and evaluation efforts for motion prediction {\citep{poiesi2015predicting,hirakawaDAPI18,ridel2018literature,quehl2017howgood} with additional datasets, probabilistic and robustness metrics, and a principled analysis of existing benchmarking practices. Furthermore, we} give an up-to-date discussion of the current state of the art and conclude with recommendations for promising directions of future research.


\section{Taxonomy}
\label{sec:taxonomy}

In this section we describe our taxonomy to decompose the motion prediction problem based on the modeling approach and the type of contextual cues, see Fig.~\ref{fig:taxonomy-overview} for an overview. 
In Sec.~\ref{sec:taxonomy:modeling_approach} and \ref{sec:taxonomy:cues} we detail the categories and give representative papers as examples of each category, and in Sec.~\ref{sec:taxonomy:classification_rules} we describe the rules for classifying the methods.


\begin{figure*}[t]
	\centering
    \includegraphics[width=0.95\linewidth,keepaspectratio]{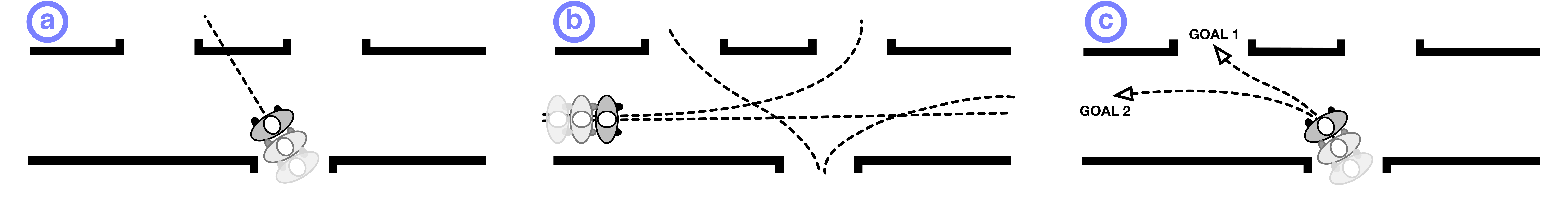}
	\caption{Illustration of the basic working principle of the modeling approaches: {\bf (a)} physics-based methods project the motion state of the agent using explicit dynamical models based on Newton's law of motion. {\bf (b)} pattern-based methods learn prototypical trajectories from observed agent behavior to predict future motion. {\bf (c)} planning-based methods include some form of reasoning about the likely goals and compute possible paths to reach those goals. In order to incorporate internal and external stimuli that influence motion behavior, approaches can be extended to account for different contextual cues. 
	}
	\label{fig:modelingapproaches}
\end{figure*}


\subsection{Modeling Approach} 
\label{sec:taxonomy:modeling_approach}

The motion modeling category subdivides the prediction approaches based on how they represent human motion and formulate the causes thereof. {\em Physics-based methods} define an explicit dynamical model based on Newton's law of motion. {\em Pattern-based methods} learn motion patterns from data of observed agent trajectories. {\em Planning-based methods} reason on motion intent of rational agents (see Fig.~\ref{fig:modelingapproaches}). The categorization can be seen to differ also in the level of cognition typically involved in the prediction process: physics-based methods follow a reactive sense-predict scheme, pattern-based methods follow a sense-learn-predict scheme, and planning-based methods follow a sense-reason-predict scheme in which agents reason about intentions and possible ways to the goal.

\begin{enumerate}[label*=\arabic*.]
	\item{\textbf{Physics-based methods} (Sense -- Predict): motion is predicted by forward simulating a set of explicitly defined dynamics equations that follow a physics-inspired model. Based on the complexity of the model, we recognize the following subclasses:
	\begin{enumerate}[label*=\arabic*.]
		\item \textbf{Single-model methods} define a single dynamical motion model, e.g. \citep{elnagarISCIRA2001,zernetsch2016trajectory,Luber2010,coscia2018long,pellegrini2009you,yamaguchiCVPR2011,aoude2010threat,petrich2013map} 
		\item \textbf{Multi-model methods} include a fixed or on-line adaptive set of multiple dynamics models and a mechanism to fuse or select the individual models, e.g.
\citep{agamennoni2012estimation,pool2017iv,kooij2018ijcv,kaempchen2004imm,althoff2008reachability,gindele2010probabilistic} 
	\end{enumerate}
	}
	\item{\textbf{Pattern-based methods} (Sense -- Learn -- Predict) approximate an arbitrary dynamics function from training data. These approaches are able to discover statistical behavioral patterns in the observed motion trajectories and are separated into two categories:
	\begin{enumerate}[label*=\arabic*.]
		\item \textbf{Sequential methods} learn conditional models over time
		and recursively apply learned transition functions for inference,
		e.g. \citep{kruse1998camera,kucner2017enabling,liao2003voronoi,aoude2011mobile,keller2014tits,vemula2017modeling,alahi2016social,GoldhammerICPR2014}
		\item \textbf{Non-sequential methods}
		directly model the distribution over full trajectories without temporal factorization of the dynamics,
		e.g. \citep{bennewitz2005learning,xiao2015unsupervised,keller2014tits,tay2008modelling,trautman2010unfreezing,kafer2010recognition,luberIROS2012} 
	\end{enumerate}
	}
	\item{\textbf{Planning-based methods} (Sense -- Reason -- Predict) explicitly reason about the agent's long-term motion goals and compute policies or path hypotheses that enable an agent to reach those goals. We classify the planning-based approaches into two categories:
	\begin{enumerate}[label*=\arabic*.]
		\item {\bf Forward planning methods} make an explicit assumption regarding the optimality criteria of an agent's motion, using a pre-defined reward function,
		e.g. \citep{vasquez2016novel, xieICCV2013, karasev2016intent, yiTRIP2016, Rudenko2017workshop, galceran2015multipolicy, best2015bayesian, bruce2004better, rosmann2017online} 
		\item {\bf Inverse planning methods} estimate the reward function or action model from observed trajectories using statistical learning techniques, e.g. \citep{ziebart2009planning,kitani2012activity,rehder2017pedestrian,kuderer2012feature, pfeiffer2016predicting, chung2012incremental, shenTransferable2018, lee2017desire, walker2014patch, huang2016deep}
	\end{enumerate}
	}
\end{enumerate}

Figure~\ref{fig:paperstatistics} shows the publications trends over the last years, color-coded by modeling approach. The number of related works is strongly increasing during the last two years in particular for the pattern-based methods.

\subsection{Contextual Cues}
\label{sec:taxonomy:cues}

\begin{figure}[t!]
	\centering
	\includegraphics[width=0.99\columnwidth]{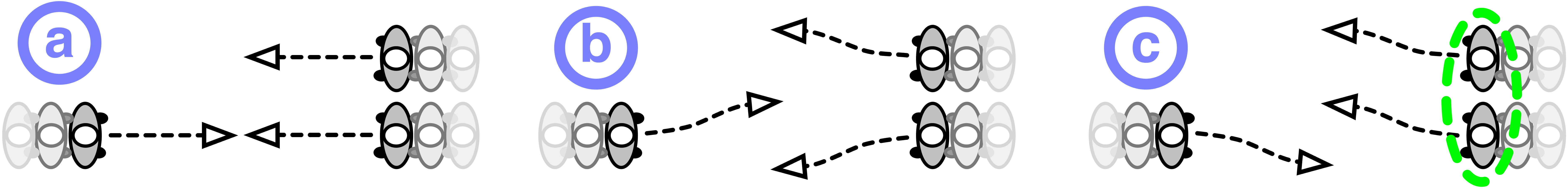}
	\caption{Dynamic environment cues: {\bf (a)} unaware, {\bf (b)} individual-aware, {\bf (c)} group-aware (accounting for social grouping cues, in green).}
	\label{fig:dynamiccontextualcues}
\end{figure}

\begin{figure}[t]
	\centering
	\includegraphics[width=0.72\columnwidth]{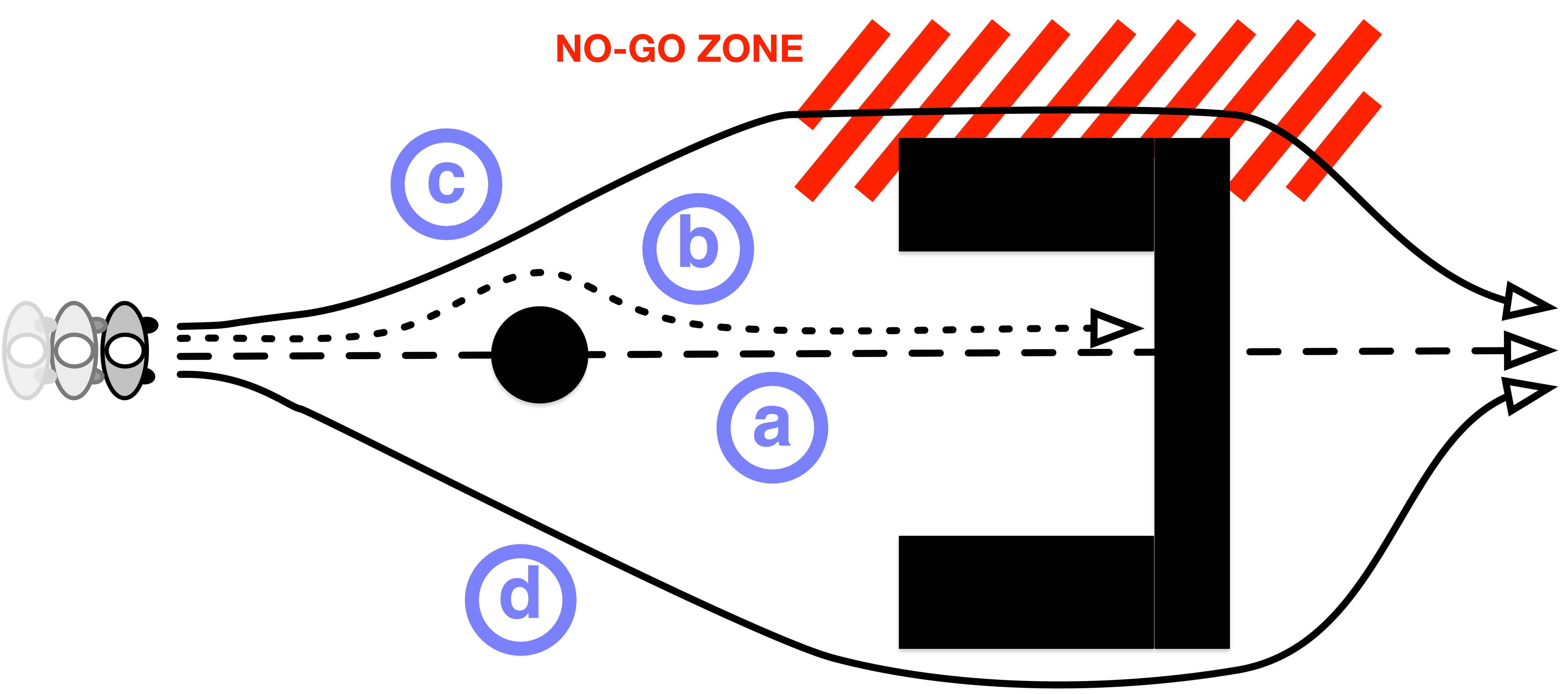}
	\caption{Static environment cues: {\bf (a)} unaware (ignoring any static objects, dashed line), {\bf (b)} obstacle-aware (accounting for unmodeled obstacles, dotted line), {\bf (c)} map-aware (accounting for a topometric environment model avoiding local minima, solid line), {\bf (d)} semantics-aware (solid line).}
	\label{fig:staticcontextualcues}
\end{figure}

We define contextual cues to be all relevant internal and external stimuli that influence motion behavior and categorize them based on their relation to the target agent, other agents in the scene and properties of the static environment, see Fig.~\ref{fig:dynamiccontextualcues} and Fig.~\ref{fig:staticcontextualcues}.
\begin{enumerate}[label*=\arabic*.]
	\item Cues of the \textbf{target agent} include
	\begin{enumerate}[label*=\arabic*.]
		\item \textbf{Motion state} (position and possibly velocity), e.g. \citep{ferrer2014behavior, elfring2014learning, pellegrini2009you, kitani2012activity, karasev2016intent, ziebart2009planning, kooij2018ijcv, trautman2010unfreezing, kuderer2012feature, bennewitz2005learning, kucner2017enabling,bera2016glmp}
		\item \textbf{Articulated pose} such as head orientation \citep{unhelkar2015human,kooij2014eccv,kooij2018ijcv,roth2016iv,hasan2018mx} or full-body pose \citep{quintero2014pedestrian,minguez2018pedestrian}
		\item \textbf{Semantic attributes} such as the age and gender \citep{ma2016forecasting}, personality \citep{bera2017aggressive}, and awareness of the robot's presence \citep{oli2013human,kooij2018ijcv} 
	\end{enumerate}
	\item With respect to the \textbf{dynamic environment} we distinguish
	\begin{enumerate}[label*=\arabic*.]
		\item \textbf{Unaware methods}, which compute motion predictions for the target agent not considering the presence of other agents, e.g. \citep{zhu1991hidden,elnagarTSMC1998,elnagarISCIRA2001,bennewitz2005learning,thompson2009probabilistic,kim2011gaussian,wang2016building,kucner2013conditional,bennewitz2005learning,thompson2009probabilistic,kim2011gaussian,wang2016building,kucner2013conditional}
		\item \textbf{Individual-aware methods}, which account for the presence of other agents, e.g. \citep{Luber2010,elfring2014learning,ferrer2014behavior,kooij2018ijcv,trautman2010unfreezing,vemula2017modeling,kuderer2012feature,alahi2016social}
		\item \textbf{Group-aware methods}, which account for the presence of other agents as well as social grouping information. This allows to consider agents in groups, formations or convoys that move differently than independent agents, e.g. \citep{yamaguchiCVPR2011,pellegrini2010improving,robicquet2016learning,singh2009modelling,qiu2010modeling,karamouzas2012simulating,seitz2014pedestrian}
	\end{enumerate}
	\item With respect to the \textbf{static environment} we distinguish
	\begin{enumerate}[label*=\arabic*.]
		\item \textbf{Unaware methods}, which assume an open-space environment, e.g. \citep{Foka2010, schneider2013gcpr, kruse1998camera, bennewitz2002using, ellis2009modelling,jacobsRAL2017, vasquez2008intentional, unhelkar2015human, ferguson2015real, luberIROS2012}
		\item \textbf{Obstacle-aware methods}, which account for the presence of individual static obstacles, e.g. \citep{rehder2015goal, trautman2010unfreezing, bera2016glmp, althoff2008stochastic, vemula2017modeling, alahi2016social, elfring2014learning, ferrer2014behavior}
		\item \textbf{Map-aware methods}, which account for environment geometry and topology, e.g. \citep{ziebart2009planning, vasquez2016novel, pfeiffer2016predicting, chen2017decentralized, pool2017iv,  Rudenko2017workshop, Rudenko2018iros, kooij2018ijcv, henry2010learning, ikeda2013modeling, liao2003voronoi, chung2010mobile, yen2008goal, chung2012incremental, gong2011multi, rosmann2017online}
		\item \textbf{Semantics-aware methods}, which additionally account for environment semantics or affordances such as no-go-zones, crosswalks, sidewalks, or traffic lights, e.g. \citep{karasev2016intent, kitani2012activity, ballan2016knowledge, ma2016forecasting, zheng2016generating, rehder2017pedestrian, coscia2018long, lee2017desire, kuhnt2016understanding}
	\end{enumerate}
\end{enumerate}

In the following Sections \ref{sec:posterior_distribution:physics_based},~\ref{sec:posterior_distribution:motion_patterns} and \ref{sec:posterior_distribution:planning_based} we survey the different classes of the motion model category.
We detail contextual cues categories in Section \ref{sec:classification_contextualcues}. In each section we discuss methods in the order of increasing complexity, considering inheritance of ideas and grouped by the similarity of the motion modeling techniques.

\subsection{Classification Rules}
\label{sec:taxonomy:classification_rules}

Some of the surveyed papers may not fall univocally into a single class of our taxonomy, especially those using a mixture of different approaches, e.g. the work by \cite{bennewitz2005learning} which combines a non-sequential clustering approach with sequential HMM inference. For those borderline cases, we adopt the following rules:
\\
\emph{i)} We classify methods primarily in the category that best describes the modelling approach over the inference method, e.g. for \citep{bennewitz2005learning} we give more weight to the clustering technique used for modelling the usual human motion behavior. 
\\
\emph{ii)} Some approaches add sub-components from other categories in their main modeling approach, e.g. planning-based approaches using physics-based transition functions \citep{van2008interactive, Rudenko2018icra}, physics-based methods tuned with learned parameters \citep{ferrer2014behavior}, planning-based approaches using inverse reinforcement learning to recover the hidden reward function of human behaviors \citep{ziebart2009planning, kitani2012activity}. We classify such approaches based on their main modeling method.
\\
\emph{iii)} Methods that use behavior cloning (imitation of human behaviors with supervised learning techniques), i.e. learn/recover the motion model directly from data, are classified as pattern-based approaches \citep{schmerling2017multimodal, zheng2016generating}. 
In contrast to that, imitation learning techniques that reason on policies (e.g. using generative adversarial imitation learning \citep{li2017inferring}) are classified as planning-based methods.

Furthermore, a single work is categorized into three contextual cues' classes with respect to its perception of the target agent, static and dynamic contextual cues.

\section{Physics-based Approaches}
\label{sec:posterior_distribution:physics_based}

Physics-based models generate future human motion considering a hand-crafted, explicit dynamical model $f$ based on Newton's laws of motion. 
A common form for $f$ is $\dot{\mathbf{\state}}_t = f(\mathbf{\state}_t, \mathbf{\action}_t, t) + w_t$ where $\mathbf{\action}_t$ is the (unknown) control input and $w_t$ the process noise. 
In fact, motion prediction can be seen as inferring $\mathbf{\state}_t$ and $\mathbf{\action}_t$ from various estimated or observed cues.

A large variety of physics-based models have been developed in the target tracking and automatic control communities to describe motion of dynamic objects in ground, marine, airborne or space applications, typically used as building blocks of a recursive Bayesian filter or multiple-model algorithm. These models differ in the type of motion they describe such as maneuvering or non-maneuvering motion in 2D or 3D, and in the complexity of the target's kinematic or dynamic model and the complexity of the noise model. See \citep{rongliTAES03survey, rongliTAES10survey} for a survey on physics-based motion models for target tracking.

We subdivide physics-based models into (1) \emph{single-model approaches} that rely on a single dynamical model $f$ and (2) \emph{multi-model approaches} that {involve} several modes of dynamics (see Fig.~\ref{fig:pictograms-physics-based}).

\subsection{Single-model Approaches}

\subsubsection{Early works and basic models}
Many approaches to human motion prediction represent the motion state of target agents as position, velocity and acceleration and use different physics-based models for prediction. Among the simplest ones are kinematic models 
without considering forces that govern the motion. Popular examples include the constant velocity model (CV) that assumes piecewise constant velocity with white noise acceleration, the constant acceleration model (CA) that assumes piecewise constant acceleration with white noise jerk, the coordinated turn model (CT) that assumes constant turn rate and speed with white noise linear and white noise turn acceleration or the more general curvilinear motion model by \cite{bestTAES97}. The bicycle model is an often used {as an approximation to model the vehicle dynamics} 
(see e.g. \citep{schubertFUSION08}).

A large number of works across all application domains rely on kinematic models for their simplicity and acceptable performance under mild conditions such as tracking with little motion uncertainty and short prediction horizons. Examples include \citep{mogelmose2015trajectory} for hazard inference from linear motion predictions of pedestrians or \citep{elnagarISCIRA2001} for Kalman filter-based (KF) prediction of dynamic obstacles using a constant acceleration model. \cite{barth2008will} use the coordinated turn model for one-step ahead prediction in an Extended Kalman Filter (EKF) to track oncoming vehicles from point clouds generated by an in-car stereo camera. \cite{batz2009recognition} use a variant of the coordinated turn model for one-step motion prediction of vehicles within an Unscented KF to detect dangerous situations based on predicted mutual distances between vehicles.

Dynamic models account for forces which, following Newton's laws, are the key descriptor of motion. Such models can become complex when they describe the physics of wheels, gearboxes, engines, or friction effects. In addition to their complexity, forces that govern the motion of other agents are not directly observable from sensory data. This makes dynamic models more challenging for motion prediction. \cite{zernetsch2016trajectory} use a dynamic model for trajectory prediction of cyclists that contains the driving force and the resistance forces from acceleration, inclination, rolling and air. The authors show experimentally that long-term predictions up to 2.5 sec ahead are geometrically more accurate when compared to a standard CV model.

Autoregressive models (ARM) that, unlike first-order Markov models, account for the history of states have also been used for motion prediction. \cite{elnagarTSMC1998} employ a third-order ARM to predict the next position and orientation of moving obstacles using maximum-likelihood estimation of the ARM parameters. \cite{caiECCV06} use a second-order ARM for single step motion prediction within a particle filter for visual target tracking of hockey players. The early work by \cite{zhu1991hidden} uses an autoregressive moving average model as transition function of a Hidden Markov Model (HMM) to predict occupancy probabilities of moving obstacles over multiple time steps with applications to predictive planning.

Physics-based models are used for motion prediction by recursively applying the dynamics model $f$ to the current state of the target agent. So far, with the exception of \citep{zhu1991hidden}, the works described above make only one-step ahead predictions and ignore contextual cues from the environment. To account for context, the dynamics model $f$ can be extended by additional forces, model parameters or state constraints as discussed hereafter. 

\begin{figure*}[t!]
	\centering		\includegraphics[width=0.99\linewidth,keepaspectratio]{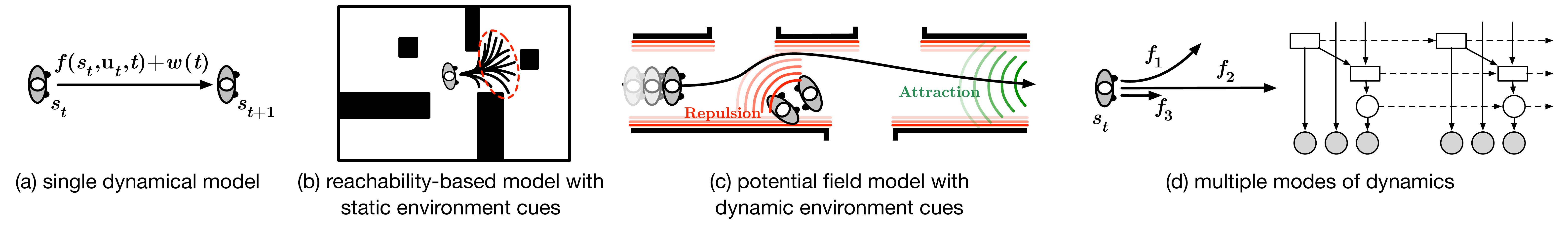}
	\caption{Examples of the physics-based approaches: {\bf (a)} a method with a single dynamical model, {\bf (b)} a reachability-based method, which accounts for all possible transitions from the given motion state, {\bf (c)} an attraction-repulsion approach, which accounts for dynamic environment cues, {\bf (d)} a multi-model method with several modes of dynamics and the DBN switching mechanism.}
	\label{fig:pictograms-physics-based}
\end{figure*}
 
\subsubsection{Models with map-based contextual cues}
A number of approaches extend physics-based models to account for information from a map, particularly for the task of tracking ground vehicles on roads. The methods developed to this end differ in how road  constraints are derived and incorporated into the state estimation problem, see the survey by \cite{simonCTA10}. \cite{yangJAIF08}, for example, use a regular KF and project the unconstrainted state estimate onto the constrained surface for tracking on-road ground vehicles with a surveillance radar. \cite{yangFUSION05} use the technique to reduce the system model parametrization to the constrained surface. They reduce vehicle motion to a 1D curvilinear road representation for filtering. \cite{batkovicARXIV18} predict pedestrian motion along a graph with straight line edges centered on side- and crosswalks. Using a unicycle model and a control approach to keep the predictions along the edges, they evaluate long-term predictions up to 10 sec ahead. When there are several possible turns at a node, i.e. at bifurcations, predictions are propagated along all outgoing edges. Another class of techniques uses the road information as pseudo measurements, pursued e.g. by \cite{petrich2013map} who use a kinematic bicycle model for $f$ and pseudo measurements from the centerlines of lanes to predict future vehicle trajectories several seconds ahead. When there are several possible turns, e.g. at intersections, the approach generates new motion hypothesis for each relevant lane by using an EKF. 

When agents move freely, e.g. do not comply with road constraints, we need different ways to represent free space and account for map information. To this end, several authors propose grid-based \citep{luberIJRR11,rehder2015goal,coscia2018long} and more general graph-based space discretizations \citep{aoude2010threat,koschiset}.
\cite{luberIJRR11} use 2D laser data to track people from a mobile robot and learn a so called spatial affordance map, a grid-based spatial Poisson process from which a walkable area map of the environment can be derived. They predict future trajectories of people during lengthy occlusion events using an auxiliary PF with look-ahead particles obtained by forward-simulation of the curvilinear motion model proposed by \cite{bestTAES97}. This way, long-term predictions (up to 50 steps ahead) stay focused on high-probability regions with the result of improved tracking performance.  
\cite{rehder2015goal} also choose a regular grid to represent the belief about pedestrian locations in a linear road scenario. They propose a variant of a Bayesian histogram filter to achieve map-aware predictions 3 seconds ahead by combining forward propagation of an unicycle pedestrian model from the start and in backward direction from the goal with prior place-dependent knowledge of motion learned from previously observed trajectories. Similarly, \cite{coscia2018long} use polars grids, centered at the currently predicted agent position to represent four different local influences: a CV motion model, prior motion knowledge learned from data, semantic map annotations like ``road'' or ``grass'' and direction to goal. The next velocity is then obtained from the normalized product of the four polar distributions and forward propagated for long-term prediction of pedestrians and cyclists in urban scenarios. Like \citep{rehder2015goal}, no planning is involved and the learned prior knowledge is place-dependent. 
\cite{koschiset} exploit information on road segments connectivity and semantic regions to compute reachability-based predictions of pedestrians, similarly to \citep{rehder2015goal}. The authors formalize several relevant traffic rules, e.g. pedestrian crossing permission on the green light, as additional motion constraints. \cite{aoude2010threat} grow a tree of future trajectories for each target agent using a closed-loop RRT algorithm that samples the controls of a bicycle motion model \citep{kuwataTCST09} avoiding obstacles in the map. Based on agent's recognized intentions using an SVM classifier and features from observed trajectories, they bias the tree growth towards areas that are more likely for the agent to enter and determine the best evasive maneuver for the ego-vehicle to minimize threat at intersection scenarios. 
A reachibility-based model, such as \citep{rehder2015goal,koschiset,aoude2010threat}, is illustrated in Fig.~\ref{fig:pictograms-physics-based} (b).

So far, we discussed extensions to physics-based motion models that embed different types of map information. All those works, however, consider only a single target agent and neglect local interactions between multiple agents. Hereafter, we will discuss methods that add social situation awareness, predicting several target agents jointly.

\subsubsection{Models with dynamic environment cues}
There are several ways to incorporate local agent interaction models into physics-based approaches for prediction, one popular example being the social force (SF) model by \cite{helbing1995social}, see Fig.~\ref{fig:pictograms-physics-based} (c). Developed for the purpose of crowd analysis and egress research, the model superimposes attractive forces from a goal with repulsive forces from other agents and obstacles. Several works extend the dynamics model $f$ to include social forces e.g. for improved short-term prediction for pedestrian tracking in 2D laser data \citep{Luber2010} or image data \citep{pellegrini2009you}. 

\cite{elfring2014learning} combine the HMM-based goal estimation method introduced by \cite{vasquez2008intentional} with the basic SF-based human motion prediction by \cite{Luber2010}. For intention estimation, the observed people trajectories are summarized in a sparse topological map of the environment. Each node of the map encodes a state--destination pair, and the goal inference using the observed trajectory is carried out in a maximum-likelihood manner. \cite{ferrer2014behavior} estimate the interaction parameters of the SF for each two people in the scene individually. For this purpose several \emph{behaviors} (i.e. sets of SF parameters) are learned offline, and the observed interaction between any two people is associated to the closest ``behavior''. The approach by \cite{oli2013human} defines the robot operating in social spaces as an interacting agent, affected by the social forces. Each human is flagged as either aware or unaware of the robot, which defines the repulsive force the robot exerts on that person. Such awareness is inferred using visual cues (gaze direction and past trajectory).

In order to achieve more realistic behaviors, several extensions to the social force model are proposed. \cite{yan2014modeling} present a model that embeds social relationships in the linear combination of predefined basic social effects (attraction, repulsion and non-interaction). The motion predictor maintains several hypothesis over the social modes, in which the pedestrians are involved.
Predictive collision avoidance behavior of the SF agents is introduced by \cite{karamouzas2009predictive} and \cite{zanlungo2011social}. In particular, \cite{karamouzas2009predictive} models each agent to adapt their route as early as possible, trying to minimize the amount of interactions with others and the energy required to solve these interactions. To this end an evasion force, that depends on the predicted point of collision and the distance to it, is applied to each agent. Updates to the SF model to consider also group motion are proposed by \cite{moussaid2010walking} and \cite{farina2017walking}. 

Other agent interaction models, not based on the social forces, for example for road vehicles, have also been used. An interactive kinematic motion model for vehicles on a single lane has been proposed by \cite{treiber2000congested} to predict the longitudinal motion of a target vehicle in the presence of preceding vehicles. The model, called Intelligent Driver Model (IDM), was used e.g. by \cite{liebnerITSM13} for driver intent inference at urban intersections. 
\cite{hoermannIV17} learn the driving style of preceding vehicles by on-line estimating the IDM parameters using particle filtering and near- and far-range radar observations. Prediction of longitudinal motion of preceding vehicles, in the experiments up to 10 seconds ahead, is then obtained by forward propagation of the model. 

Several approaches exploit the \emph{reciprocal velocity obstacles} (RVO) model \citep{van2008reciprocal} for jointly predicting human motions. \cite{kim2015brvo} use the Ensemble Kalman filtering technique together with the Expectation-Maximization algorithm to estimate and improve the human motion model (i.e. RVO parameters). \cite{bera2016glmp} propose a method that dynamically estimates parameters of the RVO function for each pedestrian, moving in a crowd, namely current and preferred velocities per agent and global motion characteristics such as entry points and movement features. A follow-up work \citep{bera2017aggressive} also introduces online estimation of personality traits. Each pedestrian's behavior is characterized as a weighted combination of six personality traits (aggressive, assertive, shy, active, tense and impulsive) based on the observations, thus defining parameters of the RVO model for this person.

Other approaches instead compute joint motion predictions based on the time of possible collision between pairs of agents.
\cite{paris2007pedestrian} propose a method for modeling predictive collision avoidance behavior in simulated scenarios. For each pedestrian current velocities of their neighbors are extrapolated in the 3D $(x,y,t)$ space, and all actions that result in collision with dynamic and static obstacles are excluded. A similar problem is addressed by \cite{pettre2009experiment}, who evaluate real people trajectories in an interactive experiment and design a predictive collision avoidance approach, capable of reproducing realistic joint maneuvers, such as giving way and passing first.

Other methods propose to compute joint motion prediction based on the expected point of closest approach between pedestrians. \cite{pellegrini2009you} is the first to propose such approach called \emph{Linear Trajectory Avoidance} (LTA): the method firstly computes the expected point of closest approach between different agents, and then uses it as driving force to perform avoidance between the agents. 
Based on the LTA, \cite{yamaguchiCVPR2011} formulate a human motion prediction approach as an energy minimization problem. The energy function considers different properties 
of people motion: damping, speed, direction, attraction, being in a group, avoiding collisions. 
The approach of Yamaguchi is further improved by \cite{robicquet2016learning} by considering several different sets of the 
energy functional parameters, learned from the training data. Each set of parameters represents a distinct behavior (navigation style of the agent).

Local interaction modeling methods, as well as approaches for predicting motion in crowds, usually benefit from detecting and considering groups of people who walk together. For example, \cite{pellegrini2010improving} propose an approach to model joint trajectories of people, taking group relations into account. The proposed framework operates in two steps: first, it generates possible trajectory hypotheses for each person, then it selects the best hypothesis that maximize a likelihood function, taking into account social factors, while at the same time estimating group membership. People and relations are modeled with Conditional Random Fields (CRF). \cite{choi2010multiple} propose an interaction model that incorporates linear motion assumption, repulsion of nearby people and group coherence via synchronization of velocities. Further group motion models, e.g. \citep{singh2009modelling,qiu2010modeling,karamouzas2012simulating,seitz2014pedestrian}, developed in the simulation and visualization communities, typically address the groups cohesion with additional forces to attract members to each other, assigning leader's and follower's roles or imposing certain group formation.

A recent reachability-based pedestrian occupancy prediction method, presented by \cite{zechel2019pedestrian}, accounts both for dynamic objects and semantics of the static environment. The authors first use a physical model to determine reachable locations of a person, and then reduce the area based on the intersections with static environment and presence probabilities of other dynamic agents.
Similarly \cite{luo2019gamma} compute future agents predictions based on an optimization approach that handles physical constraints, i.e. kinematics and geometry of the agents, and behavioral constraints, i.e. intention, attention and responsibility. 

\subsection{Multi-model Approaches}
Complex agent motion is poorly described by a single dynamical model $f$. Although the incorporation of map information and influences from multiple agents render such approaches more flexible, they remain inherently limited. A common approach to modeling general motion of maneuvering targets is the definition and fusion of different prototypical motion modes, each described by a different dynamic regime $f$. Modes may be linear movements, turn maneuvers, or sudden accelerations, that over time, form sequences able to describe complex motion behavior. Since the motion modes of other agents are not directly observable, we need techniques to represent and reason about motion mode uncertainty. The primary approach to this end are multi-model (MM) methods \citep{rongliTAES05survey} and hybrid estimation \citep{hofbaur2004hybrid}. MM methods maintain a hybrid system state $\xi=(\mathbf{x},s)$ that augments the continuous valued $\mathbf{x}$ by a discrete-valued modal state $s$. Following \citep{rongliTAES05survey}, MM methods generally consist of four elements: a fixed or on-line adaptive model set, a strategy to deal with the discrete-valued uncertainties (e.g. model sequences under a Markov or semi-Markov assumption), a recursive estimation scheme to deal with the continuous valued components conditioned on the model, and a mechanism to generate the overall best estimate from a fusion or selection of the individual filters. 
For prediction, MM methods are used in several ways, to represent more complex motion, to incorporate context information from other agents and context information from the map. 
A naive MM approach, presented by \cite{pool2017iv}, predicts future motion of cyclists using a uniform mixture of five Linear Dynamic Systems (LDS) dynamics-based motion strategies: go on straight, turn $45^\circ$ or $90^\circ$ left or right. Probability of each strategy is set to zero if the predicted path does not comply with the road topology in the place of prediction.

The interactive multiple model filter (IMM) is a widely used inference technique applied on MM models with numerous applications in tracking \citep{mazor1998interacting} and predictions. For instance, \cite{kaempchen2004imm} propose a method for future vehicle states estimation that switches between constant acceleration and simplified bicycle dynamical models. Uncertainty in the next transition is explicitly modeled with Gaussian noise. \cite{schneider2013gcpr} introduce an IMM for pedestrian trajectory prediction which combines several basic motion models (constant velocity, constant acceleration and constant turn). Also \cite{schulz2015controlled} propose a method for predicting the future path of a pedestrian using an IMM framework with constant velocity, constant position and coordinated turn models. In this work, model transitions are controlled by an intention recognition system based on Latent-dynamic Conditional Random Fields: based on the features of the person's dynamics (position and velocity) and situational awareness (head orientation), intention is classified as crossing, stopping or going in the same direction. Joint vehicle trajectory estimation also using IMMs is considered by \cite{kuhnt2015towards,kuhnt2016understanding} in a method which adopts pre-defined environment geometry to estimate possible routes of each individual vehicle. Contextual interaction constraints are embedded in a Bayesian Network that estimates the evolution of the traffic situation.

Other examples of IMMs techniques are variable-structure IMM for ground vehicles \citep{kirubarajanTAES00,noeSDPST00,pannetierFUSION05,sheaSDPST00} to account for road constraints. In a recent work \cite{xie2018vehicle} combined a kinematics-based constant turn rate and acceleration model with IMM-based lane keeping and changing maneuvers mixing. The method is aware of road geometry and produces results for a varying prediction horizon.

An alternative approach to hybrid estimation problems are dynamic Bayesian networks (DBN) which inherit the broad variety of modeling schemes and large corpus of exact and approximate inference and learning techniques from probabilistic graphical models \citep{koller2009probabilistic}. An example of a DBN-based multi-model approach is given in Fig.~\ref{fig:pictograms-physics-based} (d). The seminal work of \cite{pentland1999modeling} introduces an approach to model human behaviors by coupling a set of dynamic systems (i.e. a bank of Kalman filters (KF)) with an HMM, which is a special case of the DBNs. The authors introduce a dynamic Markov system that infers human future behaviors, a set of macro-actions described by a set of KFs, based on measured dynamic quantities (i.e. acceleration, torque). The approach was used to accurately categorize human driving actions. \cite{agamennoni2012estimation} jointly model the agent dynamics and situational context using a DBN. The vehicular dynamics is described by a bicycle model whereas the context is defined by a weighted feature function to account e.g. for closeness between agents or place-dependent information from a map. The model resembles a switched Bayesian filter but considers a more general conditioning of the switch transitions and the case of multiple agents. The authors apply the model for the task of long-term multi-vehicle trajectory prediction of mining vehicles, useful for instance during GPS outages.
\cite{kooij2014eccv} propose a context-aware path prediction method for pedestrians intending to laterally cross a street, that makes use of Switching Linear Dynamical Systems (SLDS) to model maneuvering pedestrians that alternate {between} motion models (e.g. {walking straight, stopping)}. The approach adopts a Dynamic Bayesian Network (DBN) to infer the next pedestrian movements based on the SLDS model. {The latent (context) variables  relate to pedestrian awareness of an oncoming vehicle (head orientation), the distance to the curbside and the situation criticality. \cite{kooij2018ijcv} extend this work to cover a cyclist turning scenario. In another extension of \citep{kooij2014eccv}, \cite{roth2016iv} use a second context-based SLDS to model the ``braking'' and ``driving'' behaviors of the ego-vehicle}. The two SLDS sub-graphs for modeling pedestrian and vehicle paths are combined into a joint DBN, where the situation criticality latent state is shared.
\cite{gu2016motion} propose a DBN-based motion model with a particle filter inference to estimate future position, velocity and crossing intention of a pedestrian. During inference the approach considers standing, walking and running motion modes of pedestrians.
\cite{gindele2010probabilistic} is jointly modeling future trajectories of vehicles with a DBN, describing the local context of the interaction between multiple drivers with a set of numerical features. These features are used to classify the current situation of each driver and reason on available behaviors, such as ``follow'', ``sheer in'' or ``overtake'', represented as B\'ezier curves. \cite{blaiotta2019learning} also proposes a DBN for pedestrian prediction with two motion modes (walking and standing), contextual awareness flag for the oncoming vehicle and social force-based motion dynamics for pedestrians.

Techniques derived by the stochastic reachability analysis theory \citep{althoff2010reachability} form another class of hybrid approaches to compute human motion prediction. In general, those methods model agents as hybrid systems (with multiple modes) and infer agents' future motions by computing stochastic reachable sets.
The approach by \cite{althoff2008stochastic} generates the stochastic reachable sets for interacting traffic participants using Markov chains, where each chain approximates the behavior of a single agent.
Each vehicle has its own dynamics with many modes (e.g. acceleration, deceleration, standstill, speed limit), and its goal is assumed to be known.
\cite{althoff2013road} further extend \citep{althoff2008stochastic} with the over-approximative estimation of the occupancy sets. 
The method is particularly framed for hybrid dynamics (mixed discrete and continuous) where computing the exact reachability sets could be computationally unfeasible. To overcome this issue, the method proposes to intersect different occupancy sets for different abstractions of the dynamical model.
The work by \cite{bansal2019hamilton} also uses a reachability approach for solving the prediction problem for multi-models systems. The approach rather than using a probability distribution over human next actions, it uses a deterministic set of allowable human actions. This reduces the complexity of the predictor and allows for an easy certification process.
\section{Pattern-based Approaches}
\label{sec:posterior_distribution:motion_patterns}

\begin{figure*}[t!]
	\centering		\includegraphics[width=0.99\linewidth,keepaspectratio]{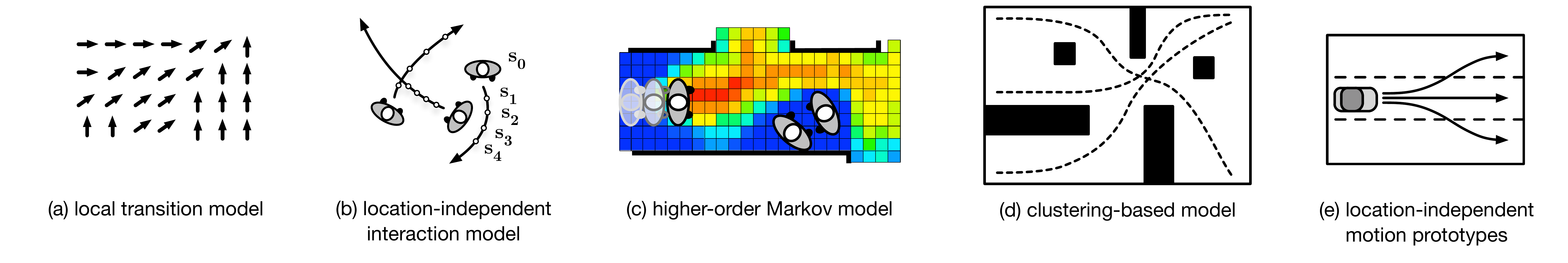}
	\caption{Examples of the pattern-based approaches: {\bf (a)} grid-based local transitions learning method, {\bf (b)} sequential location-independent transition model, which accounts for cues from the dynamic environment, {\bf (c)} higher-order sequential Markov model, {\bf (d)} clustering of full trajectories, {\bf (e)} location-independent method which learns long-term transition sequences, i.e. maneuvers.}
	\label{fig:pictograms-pattern-recognition-based}
\end{figure*}

In contrast to the physics-based approaches which 
use explicitly defined, parametrized functions of motion dynamics, 
pattern-based approaches learn the latter from data, following the \emph{Sense - Learn - Predict} paradigm. These methods learn human motion behaviors by fitting different function approximators (i.e. neural networks, hidden Markov models, Gaussian processes) to data. Many of those methods were introduced by the machine learning and computer vision communities (i.e. for behavior cloning and video surveillance applications), and later applied in robotics and autonomous navigation settings.

In our taxonomy we classify pattern-based approaches into two categories, based on the type of function approximator used:

\noindent \emph{(1) Sequential methods} typically learn conditional models, where it is assumed that the state (e.g. position, velocity) at one time instance is conditionally dependent on some sufficient statistic of the full history of past states. Many of the proposed methods are Markov models, where an $N$-th order Markov model assumes that a limited state history of $N$ time steps is a sufficient representation of the entire state history. Similarly to many physics-based approaches, sequential methods aim to learn a one-step predictor $\mathbf{\state}_{t+1} = f(\mathbf{\state}_{t-n:t} )$, where the state $\mathbf{\state}_{t+1}$ is the one step prediction and the sequence of states $\mathbf{\state}_{t-n:t}$ is the sufficient statistic of the history. In order to predict a sequence of state transitions (i.e. a trajectory), consecutive one-step predictions are made to compose a single long-term trajectory.

    

\noindent\emph{(2) Non-sequential methods} directly model the distribution over full trajectories without imposing a factorization of the dynamics (i.e. Markov assumption) as with sequential models.

\subsection{Sequential Models}
\label{sec:motion_patterns:sequential}
Sequential models are built on the assumption that the motion of intelligent agents can be described with causally conditional models over time. Similarly to the physics-based methods, transition function of sequential models has Markovian property, i.e. information on the future motion is confined in the current state of the agent.
Differently, the function, often non-parametric (e.g. Gaussian Processes, vector fields), is learned from statistical observations, and its parameters cannot be directly interpreted as for many of the physics-based methods.


\subsubsection{Local transition patterns}
Learning local motion patterns, such as probabilities of transitions between cells on a grid-map (Fig.~\ref{fig:pictograms-pattern-recognition-based} (a)), is a simple, commonly used technique for making sequential predictions 
\citep{kruse1998camera,tadokoro1993stochastic,thompson2009probabilistic,kucner2013conditional,wang2015modeling,wang2016building,ballan2016knowledge,molina2018modelling}.

Early examples of local motion patterns include the works of \cite{tadokoro1993stochastic} and \cite{kruse1998camera}.
\cite{kruse1998camera} 
build two transition models: a stochastic grid where usual motion patterns of dynamic obstacles are stored, and stochastic trajectory prediction modeled with Poisson processes.
\cite{tadokoro1993stochastic} include empirical biases to account for context features of the cells in the regions where the observations are sparse,
e.g. increasing the probability to move away from the wall, stop near a bookshelf or decrease walking speed at the crossing. More recently, \cite{thompson2009probabilistic} expand the local motion patterns model by accounting for further transitions for several steps into the future. Their method maps the motion state of the person to a series of local patches, describing where the person might be in the future. Besides the current motion state, the learned patterns are also conditioned on the final goal or the topological sub-goal in the environment. \cite{wang2015modeling} model local transition probabilities with an Input-Output HMM. Transition in each cell is conditioned both on the direction of cell entrance and the global starting point of the person's movement. \cite{jacobsRAL2017} 
use nonlinear estimation of pedestrian dynamics with the learned vector-fields to improve the linear velocity projection model. 
\cite{ballan2016knowledge} propose a Dynamic Bayesian Network method to predict not-interacting human motion based on statistical properties of human behavior. To this end a transferable navigation grid-map is learned. It encodes {functional properties} of the environment (i.e. direction and speed of the targets, crossing frequency for each patch, identification of routing points). \cite{molina2018modelling} address periodic temporal variations in the learned transition patterns, e.g. based on the time of the day.

In contrast to the discrete transition patterns discussed so far, several authors model the transition dynamics as a continuous function of the agent's motion state, using Gaussian Processes and their mixtures \citep{ellis2009modelling,joseph2011bayesian,ferguson2015real,kucner2017enabling}.
\cite{ellis2009modelling} model trajectory data in the observed environment by regressing relative motion against current position. Predictions are generated using a sequential Monte-Carlo sampling method.
\cite{joseph2011bayesian} model the multi-modal mobility patterns as a mixture of Gaussian processes with a Dirichlet process prior over mixture weights. \cite{ferguson2015real} further extends the work of \cite{joseph2011bayesian} by including a change-point detection and clustering algorithm which enables quick detection of changes in intent and on-line learning of motion patterns not seen in prior training data.
\cite{kucner2017enabling} model multimodal distributions with a Gaussian Mixture Model (GMM) in the joint velocity-orientation space.

Apart from the commonly used grid-cells, local transition patterns can be learned using a higher-level abstraction of the workspace, such as a graph of sub-goals or transition points \citep{ikeda2013modeling,han2019pedestrian}, map of connected position-velocity points \cite{kalayeh2015understanding}, Voronoi diagram \citep{liao2003voronoi}, Instantaneous Topological Map (ITM) \citep{vasquez2009incremental}, semantic-aware ITM \citep{vasishta2018building}. More flexible representation of the workspace topology is achieved this way. Combining the merits of local and global motion patterns (i.e. sequential and non-sequential models), \cite{chen2016augmented} 
model trajectories in the environment with a set of overcomplete basis vectors. 
The method breaks down trajectories into a small number of representative partial motion patterns, where each partial pattern consists of a series of local transitions. A follow-up work by \cite{habibi2018context} incorporates semantic features from the environment (relative distance to curbside and the traffic lights signals) in the learning process, improving prediction accuracy and generalization to similar environments. \cite{han2019pedestrian} propose a method to explicitly learn transition points between the local patterns.

\subsubsection{Location-independent behavioral patterns}
Unlike the local transition patterns, which are learned and applied for prediction only in a particular environment, \emph{location-independent} patterns are used for predicting transitions of an agent in the general free space \citep{aoude2011mobile,tran2014online,foka2002predictive,shalev2016long,quintero2014pedestrian} (see Fig.~\ref{fig:pictograms-pattern-recognition-based} (b)).

Several authors, e.g. \cite{foka2002predictive,shalev2016long}, use location-invariant one-step prediction as a part of collision avoidance framework using neural networks. \cite{aoude2011mobile} extend their physics-based approach \citep{aoude2010threat} by introducing location-independent GP-based motion patterns that guide the RRT-Reach to grow probabilistically weighted feasible paths of the surrounding vehicles.
\cite{tran2014online} model location-independent motion patterns of vehicles by applying spatial normalization to the trajectories in the learning set. Cartesian coordinates are turned into the relative coordinate system of the road intersection, based on the topology of the lanes.

\cite{keller2014tits} use optical flow features derived from a detected pedestrian bounding box to predict future motion. \cite{quintero2014pedestrian} instead extract full-body articulated pose. In both works, body motion dynamics {for walking and stopping} are learned using Gaussian Processes with Dynamic Model (GPDM) in a compact low-dimensional latent space. \cite{minguez2018pedestrian} extend {\citep{quintero2014pedestrian} by considering standing and starting activities as well. A first-order} HMM is used to model the transition between the activities.

Several location-independent methods learn socially-aware models of local interactions \citep{antonini2006behavioral,vemula2017modeling}. \cite{antonini2006behavioral} adapt the Discrete Choice Model from econometrics studies to predict local transitions of individuals, given the intended direction, current velocity, locations of obstacles and other people nearby.
\cite{vemula2017modeling} reformulates the non-sequential joint human motion prediction approach by \cite{trautman2010unfreezing}, discussed in Sec.~\ref{sec:motion_patterns:nonsequential}, as sequential inference with Gaussian Processes. They model the local motion of each agent conditioned on relative positions of other people in the surroundings and the person's goal.

\subsubsection{Complex long-term dependencies}
Several recent sequential methods use neural networks for time series prediction, i.e. assuming higher order Markov property \citep{sumpter2000learning,alahi2016social,bartoli2017context,varshneya2017human,sun20173dof,jain2016structural,vemula2017socialattention,GoldhammerICPR2014,schmerling2017multimodal,zheng2016generating}, see Fig.~\ref{fig:pictograms-pattern-recognition-based} (c). Such time series-based models are making a natural transition between the first order Markovian methods (e.g. local transition patterns) and non-sequential techniques (e.g. clustering-based). 
An early method, presented by \cite{sumpter2000learning} learns long-term spatio-temporal motion patterns from visual input in a known environment. The simple neural network architecture, based on natural language processing networks, quantizes partial trajectories in location/shape-space: the symbol network categorizes the object shape and locations at any time, and the context network categorizes the order in which they appear. 
\cite{GoldhammerICPR2014} learn usual human motion patterns using an ANN with the multilayer perceptron architecture. 
This method was adapted to predict motion of cyclists by \cite{zernetsch2016trajectory}.


Recurrent Neural Networks (RNN) for sequence learning, and Long Short-term Memory (LSTM) networks in particular, have recently become a widely popular modeling approach for predicting human \citep{alahi2016social, bartoli2017context,varshneya2017human,sun20173dof,vemula2017socialattention,saleh2018intent,sadeghian2018sophie}, vehicle \citep{kim2017probabilistic,altche2017lstm,park2018sequence,ding2019predicting} and cyclist \citep{pool2019context} motion.
\cite{alahi2016social} was the first one to propose a Social-LSTM model to predict joint trajectories in continuous spaces. 
Each person is modeled by an individual LSTM. 
Since humans are influenced by nearby people, 
LSTMs are connected in the social pooling system, sharing information from the hidden state of the LSTMs with the neighbouring pedestrians.
The work of \cite{bartoli2017context} extends the Social-LSTM, explicitly modeling 
human-space interactions  
by defining a ``context-aware'' pooling layer, which considers the static objects in the neighborhood of a person.
\cite{varshneya2017human}
use a Spatial Matching Network, first introduced by \cite{huang2016deep} (discussed in Sec.~\ref{subsubsection:learningplanning}), that models the spatial context of the surrounding environment, predicting the probability of the subject stepping on a particular patch. 
\cite{sun20173dof} use LSTM to learn environment- and time-specific human activity patterns in the target environment from long-term observations, i.e. covering several weeks. 
The state of the person is extended to include contextual information, i.e.
the time of the day when the person is observed. 
\cite{pfeiffer2018data} couple obstacle-awareness with an efficient representation of the surrounding dynamic agents using a 1D vector in polar angle space. \cite{bisagno2018group} add group coherence information in the social pooling layer. Saleh et al. predict  trajectories of pedestrians \citep{saleh2018intent} and cyclists \citep{saleh2018cyclist}, adapting the LSTM architecture for the perspective of a moving vehicle. Numerous other implementations of the LSTM-based predictors offer various improvements, such as increased generalizability to new and crowded environments \citep{xue2019location,shi2019pedestrian}, considering the immediate \citep{zhang2019sr} or long-term \citep{xue2017bi} intention of the agents, augmenting the state of the person with the head pose \citep{hasan2018mx} or adding a better pooling mechanism with relative importance of each person in the vicinity of the target agent \citep{xu2018encoding,fernando2018soft,pei2019human}. \cite{huynh2019trajectory} apply LSTM-based trajectory prediction in combination with local transition patterns, learned on the fly in a particular scene. Non-linear motion, historically observed in a coarse grid cell of the environment, informs the LSTM predictor.

Several  authors use LSTMs to estimate kinodynamic motion of vehicles, combining the benefits of the physics-based and the pattern-based methods \citep{raipuriaIV2017,deo2018multi}. \cite{raipuriaIV2017} augment the LSTM model with the road infrastructure indicators, expressed in the curvilinear coordinate system, to better predict motion in curved road segments. \cite{deo2018multi} propose an interaction-aware multiple-LSTM model to compute stochastic maneuver-dependent predictions of a vehicle, and augment it with an LSTM-based maneuver classification and mixing mechanism.

Other approaches use RNN as models of spatio-temporal graphs for problems that require both spatial and temporal reasoning \citep{jain2016structural,vemula2017socialattention,huang2019stgat,dai2019modeling,ivanovic2019trajectron,eiffert2019predicting}. \cite{jain2016structural} propose an approach for training sequence prediction models on arbitrary high-level spatio-temporal graphs, whose nodes and edges are represented by RNNs. The resulting graph is a feed-forward, fully differentiable, and jointly trainable RNN mixture. 
\cite{vemula2017socialattention} apply this method to jointly predict transitions in human crowds.

RNN abilities for prediction of time-series is also combined with different neural networks architectures \citep{schmerling2017multimodal, zheng2016generating, zhan2018generative,li2019coordination, choi2010multiple}.
\cite{schmerling2017multimodal} consider a traffic weaving scenario and propose a Conditional Variational Autoencoder (CVAE) with RNN subcomponents to model interactive human driver behaviors. The CVAE characterizes a multi-modal distribution over human actions at each time step conditioned on interaction history, as well as future robot action choices. \cite{zheng2016generating} describes a hierarchical policy approach that automatically reasons about both long-term and short-term goals. 
The model uses recurrent convolutional neural networks to make predictions for macro-goals (intermediate goals) and micro-actions (relative motion), which are trained independently by supervised learning, combined by an attention module, and finally jointly fine-tuned. \cite{zhan2018generative} extend this approach using Variational RNNs.
\cite{choi2019drogon} uses spatial-temporal graphs in combination with CVAE. The spatial-temporal graphs are used to model the relational influence among predicted agents. Conditions of the CVAE are represented by estimated intentions.
Also \cite{li2019coordination} propose a hierarchical architecture where an upper level (based on variational RNN) provides predictions of discrete coordination activities between agents and a lower level generates actual geometric predictions (using a Conditional Generative Adversarial Network). 
The probabilistic framework called \emph{Multiple Futures Predictor} (MFP) \citep{tang2019mfp} models joint behavior of an arbitrary number of agents via a dynamic attention-based state encoder for capturing relationships between agents, a set of stochastic, discrete latent variables per agent to allow for multimodal future behavior, as well as interactive and step-wise parallel rollouts with agent-specific RNNs to model future interactions. Furthermore, there model allows to make hypothetical rollouts under assumptions of behavior for a particular agent.


Several recent works \citep{xue2018ss,zhao2019multi,srikanth2019infer,radwan2018multimodal,vanderHeiden2019safecritic,jain2019discrete,ridel2019scene,Rhinehart_2019_ICCV} combine the benefits of sequential (e.g. RNN-based) and convolutional approaches for modeling jointly the spatial and temporal relations of the observed agents' motion. \cite{xue2018ss} introduce a hierarchical LSTM model, which combines inputs on three scales: trajectory of the person, social neighbourhood and features of the global scene layout, extracted with a CNN. \cite{zhao2019multi} propose the Multi-Agent Tensor Fusion encoding, which fuses contextual image of the environment with sequential trajectories of agents, thus retaining spatial relation between features of the environment and capturing interaction between the agents. This method is applied to both pedestrian and vehicles. Also \cite{Rhinehart_2019_ICCV} present a prediction scheme for multi-agent that combines CNNs with a generative model based on RNNs. Moreover the approach conditions the predictions on inferred intentions of the agents.
\cite{srikanth2019infer} propose a novel input representation for learning vehicle dynamics, which includes semantics images, depth information and other agents' positions. This input is projected into top-down view and fed into the autoregressive convolutional LSTM model to learn temporal dynamics.
LSTMs have been also used to predict sequence of future human movements based on a learned reward map \cite{saleh2019contextual}. 

Recently, many authors have applied the GAN architecture to achieve multi-modality in the prediction output \citep{Gupta2018SocialGAN,amirian2019social,kosaraju2019social}. For instance, \cite{Gupta2018SocialGAN} extend the Social-LSTM by using Generative Adversarial Networks and a novel variety loss which encourages the generative network to produce diverse multi-modal predictions. \cite{kosaraju2019social} use Graph Attention Network in combination with GAN architecture to better capture relative importance of surrounding agents and semantic features of the environment.

\subsection{Non-sequential Models}
\label{sec:motion_patterns:nonsequential}

Learning motion patterns in complex environments requires the model to generalize across non-uniform, context-dependent behaviors. Specifying causal constraints, e.g. through the Markovian assumption for the sequential models and additionally the particular functional form for the physics-based methods, might be too restrictive for these situations. Alternatively, instead of focusing on the local transitions of the system, \emph{non-sequential approaches} aim to directly learn a distribution over long-term trajectories, that the observed agent may follow in the future, i.e. learn a set of full motion patterns from data.

Most basic non-sequential approaches are based on clustering the observed trajectories, which creates a set of long-term motion patterns \citep{bennewitz2002using,bennewitz2005learning,chen2008pedestrian,bera2016glmp,bera2017aggressive}. This way global structure of the workspace is imposed on top of a sequential model. Clustering-based approaches are illustrated in Fig.~\ref{fig:pictograms-pattern-recognition-based} (d). \cite{bennewitz2002using, bennewitz2005learning} cluster recorded trajectories of humans into global motion patterns using the expectation maximization (EM) algorithm and build an HMM model for each cluster. For prediction, the method compares the observed track with the learned motion patterns, and reasons about which patterns best explain it. Uncertainty is handled by probabilistic mixing of the most likely patterns. Similarly, \cite{zhou2015learning} models the global motion patterns in a crowd with Linear Dynamic Systems using EM for parameters estimation. Several authors \citep{makris2002path,piciarelli2005trajectory} propose graph structures to efficiently capture the branching of trajectory clusters. \cite{chen2008pedestrian} propose a method for dynamic clustering of the observed trajectories, assuming that the set of complete motion patterns may mot be available at the time of prediction, e.g. in new environments. \cite{sung2012trajectory} propose to represent the agent's states as short trajectories rather than static positions. This higher level of abstraction provides greater flexibility to represent not only position, but also velocity and intention.
\cite{suraj2018predicting} directly use a large-scale database of observed trajectories (up to 10 millions) to estimate the future positions of a vehicle given only its position, rotation and velocity. Combining the concepts of local motion patterns and clustering, \cite{carvalho2019long} represent each cluster with a piece-wise linear vector field over an arbitrary state-space mesh.

Several approaches use Gaussian processes (GPs) or mixture models as cluster centroids representation \citep{tay2008modelling,kim2011gaussian,yoo2016visual}. \cite{tay2008modelling} introduce an approach to predict motion of a dynamic object in known scenes based on Gaussian mixture models and Gaussian processes. \cite{kim2011gaussian} model continuous dense flow fields from a sparse set of vector sequences. \cite{yoo2016visual} propose to learn most common patterns in the scene and their co-occurrence tendency using topic mixture and Gaussian mixture models. Observed trajectories are clustered into several groups of typical patterns that occur at the same time with high probability. Given a set of observed trajectories, prediction is performed considering the dominant pattern group. \cite{makansi2019overcoming} present a Mixture Density Network architecture which generates multiple hypotheses of future position in fixed interval $\Delta t$ and then fits a mixture of Gaussian or Laplace distributions to these hypothesis.

Clustering-based methods, discussed so far, generalize statistical information in a particular environment. In comparison, location-invariant methods, based on matching the observed partial trajectory to a set of prototypical trajectories, can be used in arbitrary free space \citep{hermesIVS09,keller2011dagm,xiao2015unsupervised}, see Fig.~\ref{fig:pictograms-pattern-recognition-based} (e). \cite{hermesIVS09} predict trajectories of vehicles by comparing the observed track to a set of motion patterns, clustered with a rotationally-invariant distance metric. In their Probabilistic Hierarchical Trajectory Matching (PHTM) approach, \cite{keller2011dagm} propose a probabilistic search tree of sample human trajectory snippets to find the corresponding matching sub-sequence. \cite{xiao2015unsupervised} decompose the set of sample trajectories into pre-defined motion classes, such as wandering or stopping, rotating and aligning them to start from the same point and have the longest span along the same axis. In contrast, skipping the clustering step, \cite{nikhil2018convolutional} propose a simple method to map the input trajectory of fixed length to the full future trajectory using a Convolutional Neural Network.

For interaction-aware non-sequential motion prediction, several authors consider the case with two interacting agents \citep{kafer2010recognition,luberIROS2012}. \cite{kafer2010recognition} propose a method for joint pairwise vehicle trajectory estimation at intersections. Comparing the observed motion pattern to the ones stored in a motion database, several prospective future trajectories are extracted independently for each vehicle. Probability of each pair of possible future trajectories is then estimated. \cite{luberIROS2012} model joint pairwise interactions between two people using social information. Authors learn a set of dynamic motion prototypes from observations of relative motion behavior of humans in public spaces. An unsupervised clustering technique determines the most likely future paths of two humans approaching a point of social interaction. 

In contrast to multi-agent clustering, \cite{trautman2010unfreezing} use Gaussian Processes for making single-agent trajectory predictions. Then, an interaction potential re-weights the set of trajectories based on how close people are located to each other at every moment. A follow-up work \citep{trautman2013robot} incorporates goal information into the model: the goal position is added as a training point into the GP. Another approach by \cite{su2017forecast} uses a social-aware LSTM-based crowd descriptor, which is later integrated into the deep Gaussian Process to predict a complete distribution over future trajectories of all people.

Recently, several approaches for non-sequential prediction of vehicle motion using CNNs were presented \citep{djuric2018motion,cui2019multimodal,hong2019rules}. An uncertainty-aware CNN-based vehicle motion prediction approach is presented by \cite{djuric2018motion}. Authors use a high-definition map image with projected prior motion of the target vehicle and full surrounding context as an input to the CNN, which produces the short-term trajectory of the target vehicle. The approach is extended by \cite{cui2019multimodal} to inferring multi-modal predictions. \cite{hong2019rules} propose two methods for output representation using multi-modal regression with uncertainty or stacks of grid-map crops. \cite{chai2019multipath} use a fixed set of state-sequence ``anchor'' trajectories (clustered from training data), which correspond to possible modes of future behavior, as input to a CNN for mid-level scene features inference, and predict a discrete distribution over these anchors. For each anchor, the method regresses offsets from anchor waypoints along with uncertainties, yielding a Gaussian mixture at each time step.
\section{Planning-based Approaches}
\label{sec:posterior_distribution:planning_based}

Planning-based approaches solve a sequential decision-making problem by reasoning about the future to infer a model of agent's motion. These approaches follow the \emph{Sense-Reason-Act} paradigm introduced earlier in Sec.~\ref{sec:taxonomy}. Unlike the previous two modeling approaches, the planning-based approach incorporates the concept of a rational agent when modeling human motions. By placing an assumption of rationality on the human, the models used to represent human motion must take into account the impact of current actions on the future as part of its model. As a result, much of the work covered in this section use objective functions that minimizes some notion of the total cost of a sequence of actions (motions), and not just the cost of one action in isolation.

\begin{figure}[t!]
	\centering		\includegraphics[width=0.99\linewidth,keepaspectratio]{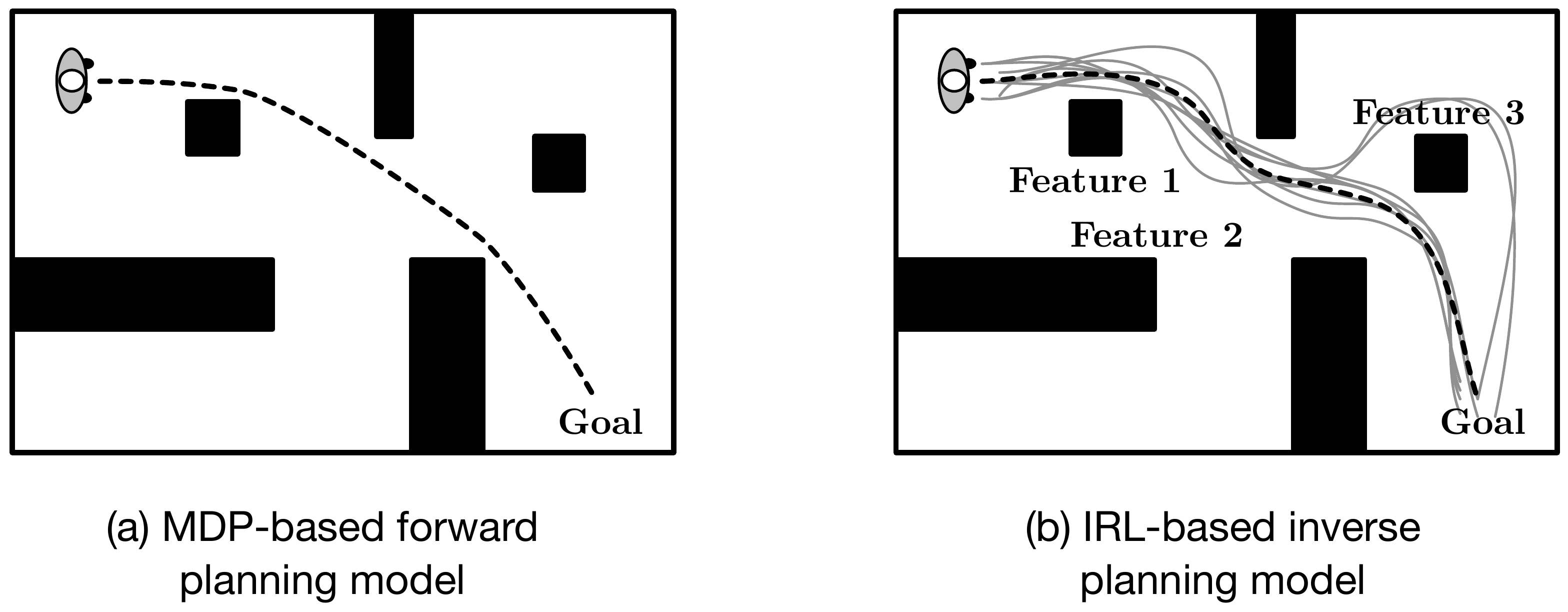}
	\caption{Examples of the planning-based approaches: {\bf (a)} forward planning approach, which uses a predefined cost function (e.g. Euclidean distance), and {\bf (b)} inverse planning approach, which infers the feature-based cost function from observations.}
	\label{fig:pictograms-planning-based}
\end{figure}

Here we classify planning-based approaches into two sub-categories, depicted in Fig.~\ref{fig:pictograms-planning-based}. \emph{Forward planning-based approaches} (Sec.~\ref{sec:forwardplanning}) use a pre-defined cost function to predict human motion, and \emph{inverse planning-based approaches} (Sec.~\ref{subsubsection:learningplanning}) infer the cost (or policy) function from observations of human behavior and then use that cost (or policy) function to predict human motion. 

\subsection{Forward Planning Approaches} 
\label{sec:forwardplanning}

\subsubsection{Motion and path planning methods}

To make basic goal-informed predictions, several methods use optimal motion and path planning techniques with a hand-crafted cost-function \citep{bruce2004better, gong2011multi,xieICCV2013,yiTRIP2016,vasishta2017natural}. \cite{bruce2004better} propose to use a path planning algorithm to infer how a person would move towards destinations in the environment. Predictions are performed using a set of learned goals. \cite{gong2011multi} use multiple long-term goal-directed path hypothesis from different homotopy classes, generated with a modified A* algorithm \citep{bhattacharya2010search}. \cite{xieICCV2013} describe a Dijkstra-based approach to predict human transitions across \emph{dark energy} fields generated from video data. Every goal location generates an attractive \emph{dark matter} Gaussian force field, while every non-walkable location generates a repulsive one. The dark matter functional objects, the map and the goals are inferred on-line using a Monte Carlo Markov Chain technique. For predicting human motion in a crowd, \cite{yiTRIP2016} introduce an energy map to model the traveling difficulty of each location in the scene, accounting for obstacles layout, moving people and stationary groups. The energy map is personalized for each observed agent, and the Fast Marching Method (FMM) \citep{sethian1996fast} is used to predict the person's path. \cite{vasishta2017natural} use A* search over the potential cost-map function for pedestrian trajectory prediction, aiming to recognize illegal crossing intention of the observed agent. The potential field accounts for semantic properties of the urban environment.

Other methods model the probabilities of future motion based on cost-to-go value estimates \citep{yen2008goal,best2015bayesian,karasev2016intent,vasquez2016novel,Rudenko2017workshop}. \cite{yen2008goal} propose a probabilistic goal-directed motion model that accounts for several goals in the environment. The method computes the cost-to-go function for each goal and evaluates the probabilities of feasible transitions in each state. A person's trajectory is predicted using a particle filter with Monte-Carlo sampling. \cite{best2015bayesian} propose a Bayesian framework that exploits the set of path hypotheses to estimate the intended destination and the future trajectory. To this end, a probabilistic dynamical model is used, which evaluates next states of the agent based on the decrease of the distance to the intended goal. Hypothesis are generated from the Probabilistic Roadmap (PRM). \cite{karasev2016intent} solve the prediction problem using a jump-Markov Decision Process, modeling the agents' behavior as switching non-linear dynamical systems. A soft MDP policy describes the nonlinear motion dynamics, and the latent goal variable governs the switches. The method uses hand-crafted costs for each surface type (e.g. sidewalk, crosswalk, road, grass), and handles time-dependent information such as traffic signals. Instead of using an MDP formulation, \cite{vasquez2016novel} proposes the Fast Marching Method (FMM) to compute the cost-to-go function for a set of goals. The predictor uses a velocity-dependent probabilistic motion model, describes the temporal evolution along the predicted path, and offers a gradient-based goal prediction that allows quick recognition of the intended destination changes.

\subsubsection{Multi-agent forward planning}

Most planning-based methods discussed so far do not consider interactions between agents in the scene. To account for presence of other agents, several authors propose to modify individual optimal policies locally with physics-based methods \citep{van2008interactive,Rudenko2018icra,wuIV18} or imitation learning \cite{muench2019composable}. A crowd simulation approach that combines global planning and local collision avoidance is presented by \cite{van2008interactive}. A global path for each agent is computed using a Probabilistic Road Map (PRM), considering only static obstacles. Local collision avoidance along the global path is done jointly for all agents using the Reciprocal Velocity Obstacles (RVO) \citep{van2008reciprocal} method. \cite{Rudenko2018icra} extend the MDP-based approaches \citep{ziebart2009planning,karasev2016intent} with a fast random-walk based method to generate joint predictions for all observed people using social forces. The authors extend their approach considering group-based social motion constraints in \citep{Rudenko2018iros}. \cite{wuIV18} extend the gridmap transition-based and reachability-based framework \citep{rehder2015goal,coscia2018long} with automatic inference of local goal points, and calculate the stochastic policy in each cell, augmenting the physics-based dynamics with optimal motion direction. The motion of pedestrians is predicted jointly with other traffic participants by risk checking of future states based on gap acceptance model \citep{brewer2006exploration}. Instead of using a physics-based approach (e.g. social forces) for augmenting the MDP-based predictor, \cite{muench2019composable} propose to learn an additional interaction-aware Q-function with imitation learning.

A number of approaches consider cooperative planning in joint state-space that includes all agents \citep{broadhurst2005monte,rosmann2015timed, bahram2016game,chen2017decentralized}. \cite{broadhurst2005monte} use Monte Carlo sampling to generate probability distributions over future trajectories of the vehicles and pedestrians jointly. The approach considers several available actions for each agent in the scene: each vehicle executes one of the hand-crafted behaviors, and humans are assumed to move freely in all directions. Also \cite{rosmann2017online} considers planning for cooperating agents. A set of topologically distinct candidate trajectories for each person is computed using trajectory optimization techniques \citep{rosmann2015timed}. Among those trajectories the best candidate is chosen according to a metric that includes group integrity, right versus left motion bias and curvature constraints. Finally, the encounter is resolved jointly in an iterative fashion. The interaction point of minimal spatial separation is computed between each two people, who adjust their trajectories accordingly, possibly switching to a different topological candidate. \cite{mavrogiannis2016decentralized} represent multi-agent interaction through the use of braid groups (topological patterns) which formalize trajectories sets. At inference time, the problem of predicting joint trajectories is posed as a graph search in a permutation graph.

Joint planning for the robot and the human is addressed by several works \citep{bandyopadhyay2013intention,galceran2015multipolicy,chen2017decentralized}. Assuming availability of a fixed set of goals, \cite{bandyopadhyay2013intention} solve an optimal motion problem for each of it, and generate appropriate motion policies. The latter are used to estimate the future evolution of the joint state-space of the robot and the human. \cite{galceran2015multipolicy} introduce a multi-policy decision-making systems to generate robot motions based on predicted movements of other agents in the scene, estimated with a changepoint-based technique \citep{fearnhead2007line}. Likelihood of future actions are sampled from the policies. The final prediction is generated by an exhaustive search of closed-loop forward simulations of these samples. The approach is well suited for predicting future macro-actions (i.e. turn left or right, slow down or speed up). 
\cite{bahram2016game} generates joint robot and agents' motions using a sequential game theory technique. The approach presents an interactive prediction and planning loop where a sequence of predictions (i.e. motion primitives) is generated for the ego-vehicle by considering the sequential evolution of the entire scene.
\cite{chen2017decentralized} develop a de-centralized multi-agent collision avoidance algorithm, which resolves local interactions with a learned joint value function that implicitly encodes cooperative behaviors.

\subsection{Inverse Planning Approaches}
\label{subsubsection:learningplanning}
Forward planning approaches, discussed so far, make an explicit assumption about the optimality criteria (reward or cost function) of an agent's motion. In this section we discuss algorithms that estimate the reward function of agents (or directly a policy) from observations, using statistical and imitation learning techniques (for a survey on imitation learning techniques applied to robotic systems we refer the reader to \citep{osa2018algorithmic}). Inverse planning methods assume that the reward or cost function, which depends on contextual and social features and defines the rational behavior, can be learned from observations (see Fig.~\ref{fig:pictograms-planning-based} (b)).

\subsubsection{Single-agent inverse planning}
In their influential work, \cite{ziebart2009planning} propose to learn a reward function yielding goal-directed behavior of pedestrians using maximum entropy inverse optimal control (MaxEnt IOC). Humans are assumed to be near-optimal decision makers with stochastic policies, learned from observations, which are used to predict motion as a probability distribution over trajectories. Building upon \citep{ziebart2009planning}, \cite{kitani2012activity} expand it to include the labeled semantic map of the environment. An IOC method takes the semantic map as an input, and learns the feature-based cost function that captures agents' preferences for e.g. walking on the sidewalk, or keeping some distance from parked cars. \cite{previtaliICMLA2016} propose an approach that adopts linear programming formulation of IRL. Using a discrete and non-uniform representation of the 2D workspace, it scales linearly with respect to the size of the environment. 
\cite{chung2010mobile} present an MDP-based model that describes spatial effects between agents and the environment. The authors use IRL to estimate cost of each state as a linear combination of trajectory length, static and dynamic obstacle avoidance and steering smoothness. Special context-based spatial effects (SSE) are identified by comparing the costs of the states, learned with IRL, and the actual observed trajectories. A follow-up work \citep{chung2012incremental} introduces a feature-based representation of SSEs, which can be modeled before being naturally observed, as in \citep{chung2010mobile}. 

Instead of IRL, other works use different techniques to learn the reward function \citep{rehder2017pedestrian,huang2016deep}. \cite{rehder2017pedestrian} solve the problem of intention recognition and trajectory prediction in one single Artificial Neural Network (ANN). The destinations and costly areas are predicted from stereo images using a recurrent Mixture Density Network (RMDN). Planning towards these destinations is performed using fully Convolutional Neural Networks (CNN). Two different architectures for planning are proposed: an MDP network and a forward-backward network, both using contextual features of the environment. \cite{huang2016deep} propose an approach that exploits two CNNs to learn a reward function considering spatial and temporal contextual information from a video sequences. A Spatial Matching Network (SMN) learns the spatial context of human motion. An Orientation Network (ON) is used to model the position variation of the object. The Dijkstra algorithm is used to find the minimum cost solution over a graph whose edges' weights are set by considering the reward function and the facing orientation computed by the two networks (SMN and ON).

All the detailed methods show that IRL or similar methods are providing powerful tools to learn human behaviors. Furthermore, \cite{shenTransferable2018} show that under some particular requirements (i.e. when the feature vector, model parameter and output representation are invariant under a rigid body transformation of the world fixed coordinate frame), IRL is suitable for learning location-independent transferable motion models.

\subsubsection{Imitation learning}

Instead of first learning a reward function and then applying planning techniques to generate motion predictions, imitation learning approaches directly extract a policy from the data. Generative Adversarial Imitation Learning (GAIL) approach, proposed by \cite{ho2016generative}, aims for matching long-term distributions over states and actions. It uses a GAN-based \citep{goodfellow2014generative} optimization procedure, in which a discriminator tries to distinguish between observations from experts and generated ones by making model rollouts. Afterwards, a model is trained to make predictions that yield similar long-term distributions over states and actions. This method has been successfully applied to learning human highway driving behavior \citep{kuefler2017imitating} and training joint pedestrian motion models \citep{Gupta2018SocialGAN}. \cite{li2017inferring} extend GAIL by introducing a component to the loss function, which maximizes the mutual information between the latent structure and observed trajectories. They test their approach in a simulated highway driving scenario, predicting the driver's actions given an input image and auxiliary information (e.g. velocity, last actions, damage), and show that it is able to imitate human driving, while automatically distinguishing between different types of behaviors.

Differently from GAIL, the deep generative technique by \cite{Rhinehart_2018_ECCV} adopts a fully differentiable model, which is easy to train without the need of an expensive policy gradient search. By minimizing a symmetrized cross-entropy between the distributions of the policy and of the demonstration data, the method allows to learn a policy that generates predictions which balance precision (i.e. avoid obstacle areas) and diversity (i.e. being multi-modal). 

\subsubsection{Multi-agent inverse planning}
In the following we review several inverse planning approaches that predict multi-agent motions 
\citep{kuderer2012feature, kretzschmar2014learning, pfeiffer2016predicting, ma2016forecasting, lee2017desire, fernando2019neighbourhood}. \cite{kuderer2012feature} and \cite{kretzschmar2014learning} propose a continuous formulation of the MaxEnt IOC \citep{ziebart2009planning} by considering a continuous spline-based trajectory representation. Their method relies on several features (e.g. travel time, collision avoidance) to capture physical and topological aspects of the pedestrians trajectories. \cite{pfeiffer2016predicting} extend the latter works by introducing the variable end-position of the each trajectory, thus reasoning over the agents' goals. \cite{walker2014patch} present an unsupervised learning approach for visual scene prediction. The approach exploits mid-level elements (i.e. image patches) as building blocks for jointly predicting positions of agents in the scene and changes in their visual appearance. The learned reward function defines the probability of a patch moving to a different location in the image. To generate predictions, the method performs a Dijkstra search on the learned reward function considering several goals. \cite{ma2016forecasting} combine the Fictitious Play \citep{brown1951iterative} game theory method with the deep learning-based visual scene analysis. Future paths hypothesis are generated jointly and iteratively: each pedestrian adapts her motion based on the predictions of the other pedestrians' actions. IRL's reward function features encode social compliance, neighborhood occupancy, distance to the goal and body orientation. Gender and age attributes, extracted with a deep network from video, define the possible average velocity of pedestrians.

\cite{lee2017desire} formulate the prediction problem as an optimization task. The method reasons on multi-modal future trajectories accounting for agent interactions, scene semantics and expected reward function, learned using a sampling-based IRL scheme. The model is wrapped into the single end-to-end trainable RNN encoder-decoder network, {called DESIRE}. The RNN architecture allows incorporation of past trajectory into the inference process, which improves prediction accuracy compared to the standard IRL-based techniques.

The previously discussed approaches for joint prediction assume multi-agent settings with rational and cooperative behavior of all agents. Differently, several approaches \citep{henry2010learning,lee2016predicting} address the problem by modeling one target person as a rational agent, acting in a dynamic environment. The influence of other agents then becomes part of the stochastic transition model of the environment. For example, \cite{henry2010learning} propose an IRL-based method for imitating human navigation in crowded environments. They conjecture that humans take into account the density and velocity of nearby people and learn a reward function that weights between these and additional features. Another approach by \cite{lee2016predicting} learns a reward function that explains behavior of a wide receiver in American football, whose strategy takes into account the behavior of the defenders. Models of the dynamic environment (e.g. linear or Gaussian Processes) are used as transitions in the IRL framework.

\cite{Rhinehart_2019_ICCV} has developed a multi-agent forecasting model called Estimating Social-forecast Probabilities (ESP) that uses exact likelihood inference (unlike VAEs or GANs) derived from a deep neural network for forecast trajectories. In contrast to most standard trajectory forecasting methods, the approach is able to reason conditionally based on additional information that it was not trained to use by accepting agent goals at test time. The approach uses a generative multi-agent model in order to perform PREdiction Conditioned On Goals (PRECOG).

\section{Contextual Cues}
\label{sec:classification_contextualcues}

In this section we discuss the categorization of the contextual cues, in those dealing with the target agent (Sec.~\ref{sec:other_classifications:target_agent}), the other dynamic agents (Sec.~\ref{sec:other_classifications:interaction}) and the static environment (Sec.~\ref{sec:other_classifications:environment_description}).
\subsection{Cues of the Target Agent}
\label{sec:other_classifications:target_agent}

Most essential cues, used to predict future states of an agent, are related to the agent itself. To this end most of the algorithms use current position and velocity of the target agent \citep{ferrer2014behavior, elfring2014learning, pellegrini2009you, kitani2012activity, karasev2016intent, ziebart2009planning, trautman2010unfreezing, kuderer2012feature, bennewitz2005learning, kucner2017enabling,bera2016glmp,wuIV18,habibi2018context,Rudenko2018iros, bahram2016game, Rhinehart_2018_ECCV, luo2019gamma,bansal2019hamilton}, often considering also the history of recent states/velocities. Position and velocity are also the main attributes of the target agent in vehicle motion prediction tasks \citep{hermesIVS09,broadhurst2005monte,kafer2010recognition}. Considering the head orientation or full articulated pose of the person \citep{quintero2014pedestrian,unhelkar2015human,kooij2014eccv,roth2016iv,schulz2015controlled,minguez2018pedestrian,kooij2018ijcv,hasan2018mx,blaiotta2019learning} 
 may bring valuable insights on the target agent's immediate intentions or their awareness of the environment. Considering additional semantic attributes of the target agent may further refine the quality of predictions: gender and age in \citep{ma2016forecasting}, 
 personality type \citep{bera2017aggressive}, class of the dynamic agent ({e.g.} a person or a cyclist in pedestrian areas, motorcycle, car or a truck on a highway) \citep{coscia2018long,ballan2016knowledge,altche2017lstm}, person's attention and awareness of the robot's presence in \citep{oli2013human,kooij2018ijcv,blaiotta2019learning}, raised arm as a bending intention indicator for cyclists \citep{pool2019context,kooij2018ijcv}.
\subsection{Cues of Other Dynamic Agents}
\label{sec:other_classifications:interaction}
Most of the time all agents navigate in a shared environment, adapting their actions, timing and route based on the others' presence and behavior. Therefore for predicting motion it is beneficial to consider interaction between moving agents. We classify the existing approaches in three categories: \emph{unaware predictors}, \emph{individual-aware predictors} and \emph{group-aware predictors}.

The class of unaware predictors includes all methods that generate motion prediction for a single agent, considering only the static contextual cues of the environment. Having no need to explicitly define or learn the interaction model, these methods are simpler to set up, require less training data to generalize, typically have less parameters to estimate. Simpler physics-based methods, such as linear velocity projection or constant acceleration models, are unaware predictors \citep{zhu1991hidden,elnagarTSMC1998,elnagarISCIRA2001,Foka2010,Bai2015,coscia2018long,koschiset,vasishta2017natural,vasishta2018building,xie2018vehicle}. Many pattern-based \citep{tadokoro1993stochastic,bennewitz2002using,bennewitz2005learning,thompson2009probabilistic,kim2011gaussian,wang2016building,kucner2013conditional,kucner2017enabling,unhelkar2015human,xiao2015unsupervised,GoldhammerICPR2014,chen2008pedestrian,chen2016augmented,suraj2018predicting,habibi2018context,hermesIVS09,molina2018modelling,kim2017probabilistic,saleh2018intent,xue2019location,xue2017bi,huynh2019trajectory,nikhil2018convolutional,sung2012trajectory,carvalho2019long,han2019pedestrian,piciarelli2005trajectory,makris2002path,makansi2019overcoming,ridel2019scene} and planning-based methods \citep{yen2008goal,ziebart2009planning, vasquez2016novel, kitani2012activity, karasev2016intent,gong2011multi,Rudenko2017workshop, Rhinehart_2018_ECCV} are unaware predictors, due to the increase of complexity for conditioning the learned transition patterns or optimal actions on the presence and positions of other agents. Methods for predicting pedestrians crossing behavior \citep{kooij2014eccv,quintero2014pedestrian,minguez2018pedestrian,roth2016iv,gu2016motion,keller2014tits,schulz2015controlled}
and cyclist motion \citep{zernetsch2016trajectory,pool2017iv,saleh2018cyclist,pool2019context} typically treat each agent individually.

Individual-aware predictors methods consider the interaction between agents by modeling or learning their influence on each other. Physics-based methods that use social forces \citep{zanlungo2011social,Luber2010,elfring2014learning,ferrer2014behavior,oli2013human,karamouzas2009predictive,blaiotta2019learning} or similar local interaction models \citep{paris2007pedestrian,pellegrini2009you,kim2011gaussian,yamaguchiCVPR2011,robicquet2016learning,pellegrini2010improving,karamouzas2010velocity,pettre2009experiment,luo2019gamma, bansal2019hamilton} are classical examples of individual-aware prediction models. A pattern-based approach by \cite{ikeda2013modeling} models deviations from the desired path using social forces. In general, however, learning joint motion patterns is a considerably harder task. For example, \cite{trautman2010unfreezing,trautman2013robot} learn unaware motion patterns, and then evaluate the predicted probability distribution over the joint paths using an explicit interaction potential. \cite{luberIROS2012} learn pairwise joint motion patterns of two humans approaching the spatial point of interaction. The approach by \cite{yoo2016visual} learns which motion patterns are likely to occur at the same time and uses this information for predicting the future motion of several dynamic objects. Some approaches propose to learn a motion policy or reward function that accounts for dynamic objects in the surrounding \citep{chung2010mobile,chung2012incremental,henry2010learning,lee2016predicting,vemula2017modeling}. \cite{Rudenko2018icra} propose an MDP planning-based method, where optimal policies of people are locally modified to account for other dynamic entities. \cite{wuIV18} and \cite{zechel2019pedestrian} discount predicted transition probabilities to states in collision with other agents. \cite{muench2019composable} decompose the interactive planning problem into two policies with the corresponding Q-functions: one for prediction in static environment, and another for interaction prediction in an obstacle-free environment. Many deep learning methods consider interactions between participants: explicitly modeling interacting entities \citep{alahi2016social,bartoli2017context,varshneya2017human,vemula2017socialattention,radwan2018multimodal,pfeiffer2018data,shi2019pedestrian,zhao2019multi,hasan2018mx,xue2018ss,su2017forecast,sadeghian2018sophie,Gupta2018SocialGAN,xu2018encoding,fernando2018soft,vanderHeiden2019safecritic,pei2019human,huang2019stgat,amirian2019social,fernando2019neighbourhood,kosaraju2019social,ivanovic2019trajectron,eiffert2019predicting,saleh2019contextual, choi2019drogon, Rhinehart_2019_ICCV}, implicitly as a result of pixel-wise prediction \citep{walker2014patch}, or by learning a joint motion policy \citep{ma2016forecasting,lee2017desire,shalev2016long,zhan2018generative}. Many vehicle prediction methods consider interaction between traffic participants, e.g. \citep{agamennoni2012estimation,kuhnt2016understanding,raipuriaIV2017,deo2018multi,kim2017probabilistic,broadhurst2005monte,kafer2010recognition, bahram2016game,srikanth2019infer,park2018sequence,djuric2018motion, cui2019multimodal, li2019coordination,hong2019rules,altche2017lstm,jain2019discrete,chai2019multipath,dai2019modeling,ding2019predicting}.  \cite{kooij2018ijcv} consider whether the ego-vehicle is on a potential collision course when predicting the road user path in their SLDS-based approach.

Group-aware predictors also recognize affiliations and relations of individual agents and respect the probability of them traveling together, as well as model an appropriate reaction of other agents to the moving group formation. For example, several physics-based methods model group relations by introducing additional attractive forces between group members \citep{yamaguchiCVPR2011,pellegrini2010improving,singh2009modelling,qiu2010modeling,karamouzas2012simulating,seitz2014pedestrian,moussaid2010walking,choi2010multiple,robicquet2016learning}. Several learning-based approaches that use LSTMs \citep{alahi2016social,bartoli2017context,varshneya2017human,pfeiffer2018data,zhang2019sr,shi2019pedestrian} may be capable of implicitly learning intra- and inter-group coherence behavior, however only the work by \cite{bisagno2018group} states this capability explicitly. A planning-based approach which implicitly respects group integrity by increasing the costs of passing between group members is presented by \cite{rosmann2017online} and an approach that explicitly models group motion constraints by \cite{Rudenko2018iros}.

Algorithms using high-level context information about dynamic agents produce more precise predictions in a variety of cases. Learning advanced social features of human motion improves interactive predictors performance, for instance different parameters for interactions of heterogeneous agents \citep{ferrer2014behavior}, advanced motion criteria such as \emph{social comfort} of navigation \citep{kuderer2012feature,luberIROS2012,pfeiffer2016predicting} or ``desire to move with the flow'' or ``avoid dense areas'' \citep{henry2010learning}. Some approaches model prior knowledge in terms of the dynamics of moving agents \citep{lee2017desire,rosmann2017online}, human attributes and personal traits \citep{ma2016forecasting}. \cite{chung2012incremental} present a general framework for learning context-related spatial effects, which affect the human motion, such as avoiding going through a waiting line, or in front of a person, who observes the work of art in a museum.

Modeling also the influence of the robot's presence on the agents' paths is another interesting line of research: \cite{trautman2010unfreezing} and \cite{oli2013human} tackle this problem by placing the robot as a peer-interacting agent among moving humans.
Several authors \citep{kuderer2012feature,kretzschmar2014learning,pfeiffer2016predicting,rosmann2017online} optimize joint trajectories for all humans and the robot. A relevant case of modeling the effect of robotic herd actions on the location and shape of the flock of animals is studied by \cite{sumpter2000learning}. 
Similarly, \cite{schmerling2017multimodal} condition human response on the candidate robot actions for modeling pairwise human-robot interaction. \cite{eiffert2019predicting} include the robot as an interacting agent in the LSTM-based predictor. \cite{tang2019mfp} compute a conditional probability density over the trajectories of other agents given the hypothetical rollout for the robot.
\subsection{Cues of the Static Environment}
\label{sec:other_classifications:environment_description}

Humans adapt their behaviors according not only to the movements of the other agents but also to the environment's shape and structure, making extensive use of its topology to reason on the possible paths to reach the long-term goal. Many existing prediction algorithms make use of such geometric information of the environment.

Some approaches produce \emph{unaware predictions}, assuming an obstacle-free environment.
This category includes several physics-based approaches \citep{zhu1991hidden,elnagarTSMC1998,elnagarISCIRA2001,Foka2010,schneider2013gcpr,Bai2015,pettre2009experiment,blaiotta2019learning}. Pattern-based methods usually model obstacles implicitly, by learning collision-free patterns \citep{tadokoro1993stochastic,kruse1998camera,bennewitz2002using,ellis2009modelling,tay2008modelling,thompson2009probabilistic,kim2011gaussian,jacobsRAL2017,vasquez2008intentional,joseph2011bayesian,ferguson2015real,wang2015modeling,wang2016building,kucner2013conditional,kucner2017enabling,sun20173dof,yoo2016visual,chen2008pedestrian,chen2016augmented,molina2018modelling,saleh2018intent,saleh2018cyclist,xue2019location,xue2017bi,hasan2018mx,huynh2019trajectory,carvalho2019long,sung2012trajectory,han2019pedestrian,piciarelli2005trajectory,makris2002path,makansi2019overcoming}. When facing a change in the obstacles' configuration, such patterns become obstacle-unaware. Location-independent motion patterns are usually obstacle-unaware \citep{luberIROS2012,hermesIVS09,xiao2015unsupervised,GoldhammerICPR2014,unhelkar2015human,nikhil2018convolutional}. Pedestrian crossing prediction methods typically assume obstacle-free environment \citep{gu2016motion,quintero2014pedestrian,roth2016iv,kooij2014eccv,schulz2015controlled,keller2014tits,minguez2018pedestrian,kooij2018ijcv}, as well as most of the vehicle prediction methods \citep{kim2017probabilistic,raipuriaIV2017,deo2018multi,suraj2018predicting,park2018sequence,altche2017lstm,ding2019predicting}, which assume the road-surface to be free of static obstacles. Finally, many methods consider only dynamic entities, but no static obstacles in the environment \citep{trautman2010unfreezing, trautman2013robot,bera2016glmp,althoff2013road,althoff2008stochastic,vemula2017modeling,alahi2016social,bartoli2017context,varshneya2017human,kim2015brvo,zanlungo2011social,kuderer2012feature,broadhurst2005monte,kafer2010recognition,vemula2017socialattention,radwan2018multimodal, bahram2016game,pfeiffer2018data,bisagno2018group,zhang2019sr,shi2019pedestrian,su2017forecast,Gupta2018SocialGAN,xu2018encoding,fernando2018soft,li2019coordination,pei2019human,huang2019stgat,amirian2019social,fernando2019neighbourhood,dai2019modeling,ivanovic2019trajectron,eiffert2019predicting}.

In several approaches the exact pose of the objects is known and utilized to compute more informed predictions (we refer to such methods as to \emph{obstacle-aware} methods).
Mainly the social force-based and similar techniques model the interaction between the moving agents and individual static obstacles \citep{van2008reciprocal,Luber2010,elfring2014learning,ferrer2014behavior,kretzschmar2014learning,pellegrini2009you,yamaguchiCVPR2011,pellegrini2010improving,robicquet2016learning,oli2013human,karasev2016intent,karamouzas2009predictive,karamouzas2010velocity,paris2007pedestrian,zechel2019pedestrian,luo2019gamma, Rhinehart_2019_ICCV}. Several location-independent pattern-based methods \citep{antonini2006behavioral,aoude2011mobile} can handle static objects avoidance.

Still, obstacle-aware methods may fail in very cluttered environments, due to the complexity of representing an environment with a set of individual obstacles. To overcome this difficulty many prediction approaches use maps which are a more complete representation of the environment (we call them \emph{map-aware} methods). Occupancy grid maps are the most common representation for these approaches, e.g. in the physics-based approach by \cite{rehder2015goal} reachability-based transitions are calculated on a binary grid-map. Particularly the planning-based approaches use this kind of representation: thanks to the map they can infer global, intentional behaviors of the agents \citep{ziebart2009planning, vasquez2016novel, pfeiffer2016predicting, xieICCV2013, previtaliICMLA2016,yiTRIP2016,chen2017decentralized,Rudenko2017workshop,Rudenko2018icra,Rudenko2018iros,henry2010learning,bruce2004better,best2015bayesian,ikeda2013modeling,liao2003voronoi,chung2010mobile,yen2008goal,chung2012incremental,gong2011multi,rosmann2017online}.
Fig.~\ref{fig:staticcontextualcues} shows the difference between the \emph{pure motion based predictions}, the \emph{obstacle-aware} and the \emph{map-aware} approaches. The latter perform better in terms of global obstacle avoidance behavior during prediction.

\emph{Semantic map based} approaches extend the map-aware approaches by considering various semantic attributes of the static environment. A semantic map \citep{karasev2016intent,kitani2012activity,rehder2017pedestrian,coscia2018long,shenTransferable2018,vasishta2017natural,vasishta2018building,Rhinehart_2018_ECCV,rhinehartArxiv2018,tadokoro1993stochastic,ballan2016knowledge,zhao2019multi,vanderHeiden2019safecritic,ridel2019scene,muench2019composable,saleh2019contextual} or extracted features from a top-down image \citep{xue2018ss,sadeghian2018sophie,kosaraju2019social,tang2019mfp} can be used to capture people preferences in walking on a particular type of surfaces. Furthermore, planning-based methods often use prior knowledge on potential goals in the environment \citep{karasev2016intent,Rudenko2017workshop,previtaliICMLA2016,vasquez2016novel,best2015bayesian}. Location- and time-specific information in the particular environment may help to improve prediction quality \citep{sun20173dof,molina2018modelling}.

Due to the high level of structure in the environment, methods in autonomous driving scenarios extensively use available semantic information, such as street layout and traffic rules \citep{kuhnt2016understanding,agamennoni2012estimation,gu2016motion,keller2014tits,lee2017desire,kooij2014eccv,petrich2013map,pool2017iv,srikanth2019infer,djuric2018motion,xie2018vehicle,cui2019multimodal,hong2019rules,jain2019discrete,chai2019multipath,pool2019context,choi2010multiple} or current state of the traffic lights \citep{karasev2016intent,gu2016motion,jain2019discrete}, also for predicting pedestrian and cyclist motion \citep{habibi2018context,koschiset,kooij2018ijcv}. 


\section{Motion Prediction Evaluation}
\label{sec:evaluation}

An important challenge for motion prediction methods is the design of experiments to evaluate their performance with respect to other methods and the requirements from the targeted application. In this section we review and discuss common metrics and datasets to this end.

\subsection{Performance Metrics}
\label{sec:evaluation:metrics}
Due to the stochastic nature of human decision making and behavior, exact prediction of trajectories is rarely possible, and we require measures to quantify the similarity between predicted and actual motion. Different prediction types -- see Fig.~\ref{introduction:fig:overview} -- require different measures: for single trajectories we need geometric measures of trajectory similarity or final displacement, for parametric and non-parametric distributions over trajectories we can use geometric measures as well as difference measures for probability distributions.  Metrics, commonly used in the literature, are summarized in Table~\ref{tab:metrics}.

\subsubsection{Geometric accuracy metrics}
\label{sec:evaluation:metrics:geometric}
Geometric measures are the most commonly used across all application domains. Several surveys have considered the topic of trajectory analysis and comparison \citep{zhang2006comparison, morris2008survey, zheng2015trajectory, quehl2017howgood, pan2016mining} where, based on the previous ones, only the recent survey by \cite{quehl2017howgood} specifically considers geometric similarity measures for trajectory prediction evaluation. In addition to that, we review the probabilistic metrics and the assessment of distributions with geometric methods in Sec.~\ref{sec:evaluation:metrics:probabilistic}, and the experiments to evaluate robustness in Sec.~\ref{sec:evaluation:metrics:other}.

Summarizing \citep{morris2008survey,quehl2017howgood}, we consider eight metrics:

{\bf Mean Euclidean Distance} (MED), also called \emph{Average Displacement Error} (ADE), averages Euclidean distances between points of the predicted trajectory and the ground truth that have the same temporal distance from their respective start points. An alternate form computes MED in a subspace between coefficients of the trajectories' principal components (PCA-Euclid). A third variant (MEDP) is a path measure able to compare paths of different length. For each $(x,y)$-point of the predicted path, the nearest ground truth point is searched. Being a path measure, MEDP is invariant to velocity differences and temporal misalignment but does not account for temporal ordering. A fourth variant (n-ADE) measures MED only on non-linear segments of trajectories. MED measures are widely used by many authors across all domains, see Table \ref{tab:metrics}. Many authors evaluate probabilistic predictions by computing expected MED under the predictive distribution, referring to it as \emph{mean ADE}, \emph{weighted mean ADE}, or, abusing notation, simply MED or ADE. This type of evaluation, however, does not measure how good the predictive distribution matches the ground truth distribution, falling short of being a true probabilistic measure. For example, it favors point predictions and avoids larger variances, as they often increase the expected ADE.

{\bf Dynamic Time Warping} (DTW) \citep{Berndt1994DTW} computes a similarity metric between trajectories of different length as the minimum total cost of warping one trajectory into another under some distance metric for point pairs. As DTW operates on full trajectories, it is susceptible to outliers.

{\bf Modified Hausdorff Distance} (MHD) \citep{dubuisson1994modified} is related to the Hausdorff distance as the maximal minimal distance between the points of predicted and actual trajectory. MHD was designed to be more robust against outliers by allowing slack during matching and to compare trajectories of different length. A further variant is the \emph{trajectory Hausdorff} measure (THAU) \citep{Lee2007TCP}, a path metric that computes a weighted sum over three distance terms each focusing on differences in perpendicular direction, length, and orientation between the paths. The weights can be chosen to be application-dependent.

{\bf Longest Common Subsequence} (LCS) \citep{Buzan2004LCSS} aligns two trajectories of different length so as to maximize the length of the common subsequence, i.e. the number of matching points between both trajectories. A good match is determined by thresholding a pair-wise distance and time difference  where not all points need to be matched. LCS is more robust to noise and outliers than DTW but finding suitable values for the two thresholds is not always easy.

{\bf CLEAR multiple object tracking accuracy} (CLEAR-MOTA) was initially introduced as a performance metric for target tracking \citep{Bernardin2008CLEARMOTA}. In the context of prediction evaluation, it is similar to LCS in that it sums up good matches between points on the predicted trajectory and the ground truth. 
The difference is that the concept of pair-wise matches/mismatches is more complex including false negatives, false positives and non-unique correspondences.

In addition to the metrics considered in \citep{morris2008survey,quehl2017howgood}, relevant metrics used in the reviewed literature include the \emph{Quaternion-based Rotationally Invariant LCS} (QRLCS), which is the rotationally invariant counterpart of LCS \citep{hermesIVS09}, and several measures that quantify different geometric aspects in addition to trajectory or path similarity:

{\bf Final Displacement Error} (FDE) measures the distance between final predicted position and the ground truth position at the corresponding time point. If the prediction is represented by a distribution, many authors compute expected FDE. FDE however, is not appropriate when there are multiple possible future positions.

{\bf Prediction Accuracy} (PA) uses a binary function to classify a prediction as correct if the predicted position fulfills some criteria, e.g. is within a threshold distance away from the ground truth. Percentage of correctly predicted trajectories is then reported. PA allows to incorporate suitable invariances into the distance function such as allowing certain types of errors. 

As also pointed out by \cite{quehl2017howgood}, the challenge in choosing a suitable measure is that each of these measures usually produce quite different results. For the sake of an unbiased and fair evaluation of different prediction algorithms, measures should be chosen not to suit a particular method but based on the requirements from the targeted application. An application which includes a lot of different velocities, for example, should not solely rely on path measures. 


\begin{table*}[ptbh]
\centering
\footnotesize
\begin{tabular}{p{1.4cm}p{3.3cm}p{11.4cm}}
\hline
 & {\bf Metric} & {\bf Used by} \\ \hline
Geometric & Average Displacement Error (ADE) & \cite{pellegrini2009you,yamaguchiCVPR2011,alahi2016social,sun20173dof,bartoli2017context,vemula2017modeling,karasev2016intent,kim2015brvo,vasquez2008intentional,yiTRIP2016,rosmann2017online,yoo2016visual,schulz2015controlled,zernetsch2016trajectory,pool2017iv,minguez2018pedestrian,wuIV18,hermesIVS09,raipuriaIV2017,deo2018multi,kim2017probabilistic,vemula2017socialattention,radwan2018multimodal,pfeiffer2018data,kooij2018ijcv,quintero2014pedestrian,saleh2018intent,saleh2018cyclist,bisagno2018group,xue2019location,zhang2019sr,shi2019pedestrian,zhao2019multi,xue2017bi,hasan2018mx,xue2018ss,su2017forecast,srikanth2019infer,sadeghian2018sophie,park2018sequence,djuric2018motion,xie2018vehicle,Gupta2018SocialGAN,huynh2019trajectory,nikhil2018convolutional,xu2018encoding,fernando2018soft,cui2019multimodal,luo2019gamma,hong2019rules,pei2019human,altche2017lstm,huang2019stgat,chai2019multipath,amirian2019social,blaiotta2019learning,dai2019modeling,kosaraju2019social,ivanovic2019trajectron,eiffert2019predicting,saleh2019contextual,choi2019drogon,Rhinehart_2018_ECCV,fernando2019neighbourhood,li2019coordination,jain2019discrete} \vspace{3pt} \\ 
 & Final Displacement Error (FDE) & \cite{varshneya2017human,alahi2016social,vemula2017modeling,chung2010mobile,vemula2017socialattention,radwan2018multimodal,bisagno2018group,xue2019location,zhang2019sr,shi2019pedestrian,zhao2019multi,xue2017bi,hasan2018mx,xue2018ss,su2017forecast,sadeghian2018sophie,Gupta2018SocialGAN,huynh2019trajectory,nikhil2018convolutional,xu2018encoding,fernando2018soft,luo2019gamma,pei2019human,huang2019stgat,amirian2019social,blaiotta2019learning,kosaraju2019social,ivanovic2019trajectron,eiffert2019predicting,choi2019drogon} \vspace{3pt} \\
 & Modified Hausdorff Distance (MHD) & \cite{vasquez2016novel,kitani2012activity,jacobsRAL2017,Rudenko2017workshop,Rudenko2018icra,Rudenko2018iros,yoo2016visual,coscia2018long,shenTransferable2018,habibi2018context,fernando2019neighbourhood,saleh2019contextual} \vspace{3pt} \\
 & Prediction Accuracy (PA) & \cite{ferrer2014behavior,ikeda2013modeling,bera2016glmp,best2015bayesian,ding2019predicting,hong2019rules} \\ \hline
Probabilistic & Negative Log Likelihood (NLL) & \cite{coscia2018long, Rudenko2017workshop, suraj2018predicting,jain2019discrete,chai2019multipath,pool2019context,makansi2019overcoming,ivanovic2019trajectron,Rhinehart_2019_ICCV} \vspace{3pt} \\
 & Negative Log Loss (NLL) & \cite{ma2016forecasting,previtaliICMLA2016, vasquez2016novel,kitani2012activity,tang2019mfp} \vspace{3pt} \\
 & Predicted Probability (PP) & \cite{kooij2014eccv,kooij2018ijcv,rehder2015goal,Rudenko2018icra,Rudenko2018iros} \vspace{3pt} \\
 & Min. Avg. or Final Displacement Error (mADE, mFDE) & \cite{lee2017desire,park2018sequence,Rhinehart_2018_ECCV,Rhinehart_2019_ICCV,ridel2019scene,ivanovic2019trajectron,amirian2019social,chai2019multipath,vanderHeiden2019safecritic,hong2019rules,li2019coordination,tang2019mfp} \vspace{3pt} \\
 & Cumulative Probability (CP) & \cite{suraj2018predicting} \\
\hline
\end{tabular}
\vspace{6pt}
\caption{Metrics to evaluate motion prediction}
\label{tab:metrics}
\end{table*}

\subsubsection{Probabilistic accuracy metrics}
\label{sec:evaluation:metrics:probabilistic}
One of the drawbacks of geometric metrics is their inability to measure uncertainty and also multimodal nature of predictions, e.g. when the target agent may take different paths to reach the goal, or when an observed partial trajectory matches several previously learned motion patterns. Moreover due to the stochasticity of the human behaviors, motion prediction algorithms need to be evaluated on their accuracy to match the underlying probability distribution of human movements. Several probabilistic accuracy metrics can be used for this purpose.

Many variational inference and machine learning algorithms \citep{mackay2003information,Bishop2006} use the Kullback-Leibler (KL) divergence \citep{kullback1951information} to measure dissimilarity of two distributions, e.g. the unknown probability distribution of human behavior $p(\mathbf{\state}_{1:T})$ and the predicted probability distribution $q(\mathbf{\state}_{1:T}|\theta)$, with $\theta$ being a set of parameters of the chosen prediction model. The KL divergence is computed as $d_\mathit{KL}(p||q) \simeq	\sum_{\mathbf{\state}_{1:T} \in \mathbb{S}} \{ - p(\mathbf{\state}_{1:T})\log{q(\mathbf{\state}_{1:T}|\theta)} + p(\mathbf{\state}_{1:T})\log{p(\mathbf{\state}_{1:T})}\}$ with the space of all trajectories $\mathbb{S}$.
Minimizing the $d_\mathit{KL}(p||q)$ corresponds to maximizing the log-likelihood function for $\theta$ under the predicted distribution $q(\mathbf{\state}_{1:T}|\theta)$. Different surveyed papers have adopted variants of the KL divergence
as accuracy metric for their stochastic predictions.

For example, the {\bf average Negative Log Likelihood} or {\bf average Negative Log Loss} evaluates the negative log likelihood term \big($\simeq \sum_{\mathbf{\state}_{1:T} \in \mathbb{D}} \log{q(\mathbf{\state}_{1:T}|\theta)}$\big) of $d_\mathit{KL}$ from a set of ground truth demonstrations $\mathbb{D} = \left\{\mathbf{\state}_{1:T}^i\right\}_{i=1}^N$ with the total number of demonstrations $N$.
Furthermore, several approaches use the {\bf Predicted Probability} (PP) metric, \big($\simeq \sum_{t=1}^T q(\mathbf{\state}_t|\theta)$\big) or its negative logarithm, to calculate the probability of the ground truth path (\emph{i.e} $\mathbf{\state}_{1:T}$) on the predicted states distribution. 
For the above metrics, the computation of the log likelihood depends on the chosen model, its induced graph and the corresponding factorization.
Finally, the {\bf Cumulative Probability} (CP) metric computes the fraction of the predictive distribution that lies within a radius $r$ from the correct position for various values of $r$.

Several recently introduced metrics follow a sampling approach to evaluate a probability distribution. {\bf Minimum Average Displacement Error} (mADE) metric \citep{walker2016uncertain,Rhinehart_2019_ICCV,scholler2019simpler, thiede2019analyzing,tang2019mfp}, as well as \emph{variety loss}, \emph{oracle}, \emph{Minimum over N}, \emph{Best-of-N}, \emph{top n\%}, or \emph{minimum Mean Squared Distance} (minMSD), computes Euclidean distance between the ground truth position of the agent $\mathbf{\state}_t^*$ at time $t$ and the closest (or the n\% closest) of the $K$ samples from the predicted probability distribution: $\min_k ||\mathbf{\state}_t^*-\mathbf{\state}_t^{k}||$. Similarly, {\bf minimum Final Displacement Error} (mFDE) evaluates only the distribution at the prediction horizon $T$. Such metrics encourage the predicted distribution to cover multiple modes of the ground truth distribution, while placing probability mass according to the mode likelihood. An evaluation of the robustness of top 1 vs. top n\% metrics by \cite{bhattacharyya2019conditional} has shown that the \emph{top n\%} metric produces more stable results.

\subsubsection{Other performance metrics}
\label{sec:evaluation:metrics:other}
Prediction accuracy is by far the primary performance indicator in the reviewed literature across approaches and application domains. 
In particular for long-term prediction methods, authors evaluate accuracy against the prediction horizon \citep{karasev2016intent, Rudenko2018icra, wuIV18, rehder2015goal, Rudenko2018iros, galceran2015multipolicy, bahram2016game, chung2010mobile, pfeiffer2016predicting, lee2016predicting,thompson2009probabilistic,jacobsRAL2017,ikeda2013modeling,vasishta2018building,keller2014tits,quintero2014pedestrian,GoldhammerICPR2014,pfeiffer2018data,sun20173dof,raipuriaIV2017,deo2018multi,radwan2018multimodal,suraj2018predicting,hermesIVS09,xu2018encoding,blaiotta2019learning,choi2019drogon}. Much fewer authors address other aspects of robustness and investigate the range of conditions under which prediction results remain stable and how they are impacted by different types of perturbations.

Experiments to explore robustness evaluate prediction accuracy as a function of various influences: the length or duration of the observed partial trajectory until prediction (addresses the question of how long the target agent needs to be observed for a good prediction) \citep{lee2017desire, kitani2012activity,radwan2018multimodal},
the size of the training dataset \citep{vasquez2009incremental,vasishta2018building,suraj2018predicting,huynh2019trajectory}, number of agents in the scene \citep{Rhinehart_2019_ICCV}, input data sampling frequency and the amount of sensor noise \citep{bera2016glmp} or amount of anomalies in the training trajectories \citep{han2019pedestrian}. Several authors report a separate accuracy measurement for the more challenging (e.g. non-linear or anomalous) part of the test set \citep{fernando2018soft,huynh2019trajectory,kooij2018ijcv}, or evaluate the model's performance on different classes of behavior, e.g. walking or stopping \citep{saleh2018intent}. Analysis of generalization, overfitting and input utilization by a neural network, presented by \cite{scholler2019simpler}, makes a good case for robustness evaluation.

Furthermore, to quantify efficiency of a prediction method, some authors relate inference time to the number of agents in the scene \citep{Rudenko2018icra, Rudenko2018iros,thompson2009probabilistic}, and only a few papers provide an analysis of their algorithms' complexity \citep{best2015bayesian, Rudenko2018iros,chen2016augmented,keller2014tits,zhao2019multi}.

\subsection{Datasets}
\label{sec:evaluation:datasets}
In order to evaluate the quality of predictions, predicted states or distributions are usually compared to the ground truth states using standard datasets of recorded motion. Availability of annotated trajectories, represented with the sequence of states or bounding boxes in the top-down view, sets prediction benchmarking datasets aside from the other popular computer vision datasets,
where the ground truth state of the agent is not available and is difficult to estimate. 

Common recording setup includes a video-camera with static top-down view of the scene, or ground-based lasers and/or depth sensors, mounted on a static or moving platform. Detected agents in each frame are labeled with unique IDs, and their positions with respect to the global world frame are given as $(x,y)$ coordinates together with the frame time-stamp $t$, i.e. (id, $t, x, y$). Often the coordinate vector is augmented with orientation and velocity information. Furthermore, social grouping information, gaze directions, motion mode or maneuver labels and other contextual cues can be provided. Apart from this specific form of labeling, further requirements to prediction benchmarking datasets include interaction between agents, varying density of agents, presence of non-convex obstacles in the environment, availability of the semantic map and long continuous observations of the agents.

In Table~\ref{tab:datasets} we review the most popular datasets, used for evaluation in the surveyed literature. Out of many datasets, used for benchmarking by different authors, we picked those used by at least two independent teams, excluding the creators of the dataset. We believe that this is a good indication of the dataset's relevance, which also supports the primary purpose of benchchmarking -- comparing performance of different methods on the same dataset. Additionally, in Table~\ref{tab:additional-datasets} we include four recent datasets, which do not meet the selection criterion, but cover valuable aspects, missing from the earlier datasets. This includes the first dataset of cyclists trajectories \citep{pool2017iv}, the first large-scale dataset of vehicles trajectories \citep{Krajewski2018HighDdataset}, the first dedicated benchmark for human trajectory prediction \citep{sadeghiankosaraju2018trajnet} and the first dataset of human motion trajectories with accurate motion capture data \citep{rudenko2019thor}.

\begin{table*}[tbh]
\centering
\footnotesize
\begin{tabular}{p{2.2cm}p{1.0cm}p{1.0cm}p{1.1cm}p{4.1cm}p{1.7cm}p{3.2cm}}
\hline
{\bf Dataset} & {\bf Location} & {\bf Agents} & {\bf Sensors} & {\bf Scene description} & {\bf Duration and tracks} & {\bf Annotations and sampling rate} \\ \hline
\textbf{ETH} \citep{pellegrini2009you} & Outdoor & People & Camera & 2 pedestrian scenes, top-down view, moderately crowded & 25 min, 650 tracks & Positions, velocities, groups, maps \linebreak @2.5 Hz \\
\multicolumn{7}{p{16.9cm}}{Used by: \cite{varshneya2017human,bera2016glmp,alahi2016social,vemula2017modeling,trautman2010unfreezing,kim2015brvo,yamaguchiCVPR2011,chung2010mobile,vemula2017socialattention,radwan2018multimodal,pfeiffer2018data,bisagno2018group,zhang2019sr,zhao2019multi,xue2018ss,sadeghian2018sophie,Gupta2018SocialGAN,huynh2019trajectory,nikhil2018convolutional,xu2018encoding, luo2019gamma,pei2019human,huang2019stgat,amirian2019social,blaiotta2019learning,kosaraju2019social,ivanovic2019trajectron}} \\ \hline
\textbf{UCY} \citep{lerner2007crowds} & Outdoor & People & Camera & 2 pedestrian scenes (sparsely populated Zara and crowded Students), top-down view & 16.5 min, over 700 tracks & Positions, gaze directions \linebreak -- \\ 
\multicolumn{7}{p{16.9cm}}{Used by: \cite{ma2016forecasting,varshneya2017human,alahi2016social,bartoli2017context,best2015bayesian,yamaguchiCVPR2011,pellegrini2010improving,vemula2017socialattention,radwan2018multimodal,bisagno2018group,zhang2019sr,zhao2019multi,hasan2018mx,xue2018ss,sadeghian2018sophie,Gupta2018SocialGAN,huynh2019trajectory,nikhil2018convolutional,xu2018encoding,vanderHeiden2019safecritic,pei2019human,huang2019stgat,luo2019gamma,amirian2019social,blaiotta2019learning,kosaraju2019social,ivanovic2019trajectron}} \\ \hline
\textbf{Stanford Drone Dataset} \citep{robicquet2016learning} & Outdoor & People, cyclists, vehicles & Camera & 8 urban scenes, $\sim$\SI{900}{\metre\squared} each, top-down view, moderately crowded & 5 hours, 20k tracks & Bounding boxes \linebreak @30 Hz \\  
\multicolumn{7}{p{16.9cm}}{Used by: \cite{varshneya2017human,jacobsRAL2017,coscia2018long,zhao2019multi,sadeghian2018sophie,vanderHeiden2019safecritic,chai2019multipath,fernando2019neighbourhood,makansi2019overcoming,eiffert2019predicting,ridel2019scene,saleh2019contextual}} \\ \hline
\textbf{NGSIM} \citep{colyar2006us,colyar2007us} & Outdoor & Vehicles & Camera network & Recording of the US Highway 101 and Interstate 80, road segment length 640 and 500 \SI{}{\metre} & 90 min & Local and global positions, velocities, lanes, vehicle type and parameters, \linebreak @10 Hz \\ 
\multicolumn{7}{p{16.9cm}}{Used by: \cite{kuefler2017imitating,deo2018multi,zhao2019multi,altche2017lstm,li2019coordination,kalayeh2015understanding,dai2019modeling,ding2019predicting,tang2019mfp}} \\ \hline
\textbf{Edinburgh} \citep{majecka2009statistical} & Outdoor & People & Camera & 1 pedestrian scene, top-down view, 12~x~16 \SI{}{\metre\squared}, varying density of people & Several months, 92k tracks & Positions \linebreak @9 Hz \\ 
\multicolumn{7}{p{16.9cm}}{Used by: \cite{previtaliICMLA2016,elfring2014learning,Rudenko2017workshop,xue2017bi,fernando2018soft,carvalho2019long}} \\ \hline
\textbf{Grand Central Station Dataset} \citep{zhou2012understanding} & Indoor & People & Camera & Recording in the crowded New York Grand Central train station & 33 minutes & Tracklets \linebreak @25 Hz \\ 
\multicolumn{7}{p{16.9cm}}{Used by: \cite{su2017forecast,xue2017bi,xue2019location,yiTRIP2016,xu2018encoding,fernando2018soft}} \\ \hline
\textbf{VIRAT} \citep{oh2011large} & Outdoor & People, cars, other vehicles & Camera & 16 urban scenes, 20--50$^\circ$ camera view angle towards the ground plane, homographies included & 25 hours & Bounding boxes, events (e.g. entering  a vehicle or using a facility) \linebreak @10,~5 and~2~Hz \\ 
\multicolumn{7}{p{16.9cm}}{Used by: \cite{previtaliICMLA2016,vasquez2016novel,kitani2012activity,walker2014patch,xieICCV2013}} \\ \hline
\textbf{KITTI} \citep{Geiger2012CVPR} & Outdoor & People, cyclists, vehicles & Velodyne, 4 cameras & Recorded around the mid-size city of Karlsruhe (Germany), in rural areas and on highways & 21 training sequences and 29 test sequences 
& 3D Positions \linebreak @10 Hz \\ 
\multicolumn{7}{p{16.9cm}}{Used by: \cite{karasev2016intent, wuIV18, Rhinehart_2018_ECCV, lee2017desire,srikanth2019infer}} \\ \hline
\textbf{Town Center Dataset} \citep{benfold2011stable} & Outdoor & People & Camera & Pedestrians moving along a moderately crowded street & 5 minutes, 230 hand labelled tracks & Bounding boxes  \linebreak @15 Hz \\ 
\multicolumn{7}{p{16.9cm}}{Used by: \cite{ma2016forecasting,xue2018ss,xue2019location,hasan2018mx}} \\
\hline
\textbf{ATC} \citep{brscic2013person} & Indoor & People & 3D range sensors & Recording in a shopping center, 900 \SI{}{\metre\squared} coverage, varying density of people & 92 days, long tracks & Positions, orientations, velocities, gaze directions, \linebreak @10-30 Hz \\ 
\multicolumn{7}{p{16.9cm}}{Used by: \cite{Rudenko2018icra,Rudenko2018iros,molina2018modelling}} \\ \hline
\textbf{Daimler Pedestrian Path Prediction Dataset} \citep{schneider2013gcpr} & Outdoor & People & Stereo camera & Recording from a moving or standing vehicle, pedestrians are crossing the street, stopping at the curb, starting to move or bending in & 68 tracks of pedestrians, 4 sec each & Positions, bounding boxes, stereo images, calibration data \linebreak @17 Hz \\ 
\multicolumn{7}{p{16.9cm}}{Used by: \cite{schulz2015controlled,saleh2018intent,saleh2019contextual}} \\ \hline
\textbf{L-CAS} \citep{yan2017online} & Indoor & People & Velodyne & Recording in a university building from a moving or stationary robot & 49 minutes & Positions, groups, Velodyne scans \linebreak @10 Hz \\ 
\multicolumn{7}{p{16.9cm}}{Used by: \cite{sun20173dof,radwan2018multimodal}} \\
\hline
\end{tabular}
\vspace{6pt}
\caption{Overview of the motion trajectories datasets}
\label{tab:datasets}
\end{table*}

\begin{table*}[ptbh]
\centering
\footnotesize
\begin{tabular}{p{2.2cm}p{1.0cm}p{1.0cm}p{1.1cm}p{4.1cm}p{1.7cm}p{3.2cm}}
\hline
{\bf Dataset} & {\bf Location} & {\bf Agents} & {\bf Sensors} & {\bf Scene description} & {\bf Duration and tracks} & {\bf Annotations and sampling rate} \\ \hline
\textbf{Tsinghua-Daimler Cyclist} \citep{pool2017iv} & Outdoor & Cyclists & Stereo camera & Recording from a moving vehicle & 134 tracks
& Positions, road topology \linebreak @5 Hz \\ 
\multicolumn{7}{p{16.9cm}}{Used by: \cite{saleh2018cyclist}} \\ \hline
\textbf{TrajNet} \citep{sadeghiankosaraju2018trajnet} & Outdoor & People & Cameras & Superset of datasets, collecting also relevant metrics and visualization tools & Superset of image-plane and world-plane datasets & Bounding boxes and tracklets, datasets recording at different frequencies \\
\multicolumn{7}{p{16.9cm}}{Used by: \cite{xue2019location}} \\
\hline
\textbf{highD Dataset} \citep{Krajewski2018HighDdataset} & Outdoor & Vehicles & Camera & 6 different highway locations near Cologne, top-down view, varying densities with light and heavy traffic & Over 110k vehicles, 447 driven hours
& Positions and additional features, e.g. THW, TTC \linebreak @25 Hz \\
\hline
\textbf{TH\"OR} \citep{rudenko2019thor} & Indoor & People & Motion capture & Human-robot navigation study in a university lab & Over 600 person and group trajectories in 60 minutes
& Positions, head orientations, gaze directions, groups, map, Velodyne scans \linebreak @100 Hz
\\  
\hline
\end{tabular}
\vspace{6pt}
\caption{Additional motion trajectories datasets}
\label{tab:additional-datasets}
\end{table*}
\section{Discussion}
\label{sec:discussion}



There has been great progress in developing advanced prediction techniques over the last years in terms of method diversity, performance and relevance to an increasing number of application scenarios. In this section, we summarize and discuss the state of the art and pose the three questions initially raised in the introduction: \emph{Are the evaluation techniques to measure prediction performance good enough and follow best practices (Q1)}? This is discussed in Sec.~\ref{sec:discussion:benchmarking} by reviewing the existing benchmarking practices including metrics, experiments and datasets. \emph{Have all prediction methods arrived on the same performance level and the choice of the modeling approach does not matter anymore (Q2)}? This is discussed in Sec.~\ref{sec:modeling_approaches_discussion} where we consider the theoretical and demonstrated ability of the different modeling approaches to solve the motion prediction problem by accounting for contextual cues from the environment and the target agent. And: \emph{Is motion prediction solved (Q3)}? This is discussed in Sec.~\ref{sec:application_areas_discussion} by revisiting the requirements from the different application scenarios. Finally, in Sec.~\ref{sec:discussion:open_challenges} we outline open challenges and future research directions.

\subsection{Benchmarking}
\label{sec:discussion:benchmarking}



Evaluating the performance of a motion prediction algorithm requires choosing appropriate testing scenarios and accuracy metrics, as well as studying the methods's robustness against various variables, such as the number of interacting agents or amount of maneuvering in the data.

Depending on the application area, the testing scenario may be an intersection, a highway, a pedestrian crossing, shared urban street with heterogeneous agents, a home environment or a crowded public space.  Existing datasets, summarized in Sec.~\ref{sec:evaluation:datasets}, cover a wide range of scenarios, e.g. indoor \citep{zhou2012understanding,brscic2013person,rudenko2019thor} and outdoor environments \citep{pellegrini2009you,lerner2007crowds,oh2011large}, pedestrian areas \citep{majecka2009statistical,benfold2011stable}, urban zones \citep{robicquet2016learning,schneider2013gcpr} and highways \citep{colyar2006us,colyar2007us,Krajewski2018HighDdataset}, and include trajectories of various agents, such as people, cyclists and vehicles. However, these datasets are usually semi-automatically annotated and therefore only provide incomplete and noisy estimation of the ground truth positions (due to annotation artifacts). Furthermore, length of the trajectories is often not sufficient for evaluation in some application domains, where long-term predictions are required. Moreover, the amount of interactions between recorded agents is often limited or disbalanced (very few agents are interacting, ergo misinterpreting such cases is not reflected in the lower benchmark scores). Finally, relevant semantic information about static (i.e. grass, crosswalks, sidewalks, streets) and dynamic (i.e. human attributes such as age, gender or group affiliation) entities is usually not recorded.

Accuracy metrics, described in Sec.~\ref{sec:evaluation:metrics}, offer a rich choice for benchmarking, ranging from computing geometric distances between points (ADE, FDE) also accounting for temporal misalignments (DTW, MHD), to probabilistic policy likelihood measures (NLL) and sampling-based distribution evaluation (mADE). For long-term forecasts made in topologically non-trivial scenarios, results are usually multi-modal and associated with uncertainty. Performance evaluation of such methods should make use of metrics that account for this, such as negative log-likelihood or log-loss derived from the KLD. Not all authors are currently using such metrics. Even for short-term prediction horizons, for which a large majority of authors use geometric metrics only (AED, FDE), probabilistic metrics are preferable as they better reflect the stochastic nature of human motion and the uncertainties involved from imperfect sensing.

Another issue of benchmarking is related to variations in exact metric formulation and different names used for the same metric, e.g. for the ADE- and likelihood-based metrics, as indicated in Sec.~\ref{sec:evaluation:metrics}. Additionally, precision is often evaluated on a single arbitrary prediction horizon. These aspects obstruct comparison of the relative precision of various methods.

Furthermore, very few authors currently address robustness as a relevant issue/topic. This is surprising as prediction needs to be robust against a variety of perturbations when deployed in real systems. Examples includes sensing and detection errors, tracking deficiencies, self-localization uncertainties or map changes.



\subsubsection{On question 1:} We conclude that \emph{Q1} is not confirmed. Despite the numerous metrics, datasets and experiment designs, used in individual works, benchmarking prediction algorithms lacks a systematic approach with common evaluation practices.

For evaluating prediction quality, researchers should opt for more complex testing scenarios (which include non-convex obstacles, long trajectories, collision avoidance maneuvers and non-trivial interactions) and the complete set of metrics (both geometric and probabilistic). It is a good practice to condition the forecast precision on various prediction horizons, observation periods and the complexity of the scene, {e.g.} defined by how many interacting agents are tracked simultaneously. Furthermore, perfect sensing, perception and tracking is not always achieved in real-life operation, and therefore algorithms' performance ideally should be investigated in realistic conditions and supported by robustness experiments, e.g. see Sec.~\ref{sec:evaluation:metrics:other}. Performing proper performance analysis would clarify application potential and effective prediction horizon of many methods.

Similar benchmarking practices should be applied to runtime evaluation. Considering efficiency on embedded CPUs of autonomous systems is important for the algorithm's design and evaluation. To prove applicability in real-life scenarios (e.g. in the pipeline with time-sensitive local and global motion planners), discussion should include formal complexity and runtime analysis, conditioned on the scene complexity and prediction horizon.

For a fair objective comparison of the prediction algorithms, developing a standard benchmark with testing scenarios and metrics is becoming a task of critical importance, e.g. given the rapid growth in published literature (see Fig.~\ref{fig:paperstatistics}). The first attempt to build such a benchmark, TrajNet, is taken by \cite{sadeghiankosaraju2018trajnet}, with the follow up, TrajNet++, to be released soon. TrajNet is based on selected trajectories from the ETH, UCY and Stanford Drone Dataset and uses the ADE and FDE evaluation metrics. We encourage more researchers to follow this example and contribute to the unification of benchmarking practices.

\subsection{Modeling Approaches}
\label{sec:modeling_approaches_discussion}
With such a wide variety of motion modeling approaches, a natural question arises: which one should be preferred? In this section we discuss the inherent strengths and limitations of different approaches' classes and the efforts to incorporate various contextual cues. This discussion continues in Sec.~\ref{sec:application_areas_discussion} with highlighting the specifics of several key tasks in the application domains.

Physics-based approaches are suitable in those situations where the effect of other agents or the static environment, 
and the agent's motion dynamics can be modeled by an explicit transition function. Many of the physics-based approaches naturally handle joint predictions 
and group coherence. 
With the choice of an appropriate transition function, physics-based approaches can be readily applied across multiple environments, without the need for training datasets (some data for parameter estimation is useful, though). 
The downside of using explicitly designed motion models is that they might not capture well the complexity of the real world. The transition functions tend to lack information regarding the ``greater picture'', both on the spatial and the temporal scale, leading to solutions that represent local minima (``dead ends''). In practice, this limits the usability of physics-based methods to short prediction horizons and relatively obstacle-free environments.
All in all, the existence of fast approximate inference, the applicability across multiple domains under mild conditions, and the interpretability make physics-based approaches a popular option for the collision avoidance of the mobile platforms ({e.g.} self-driving vehicles, service robots) and the people tracking applications. 


Pattern-based approaches are suitable for environments with complex unknown dynamics (e.g. public areas with rich semantics), and can cope with comparatively large prediction horizons. However, this requires ample data that must be collected for training purposes in a particular type of location or scenario. One further issue is the generalization capability of such learned model, whether it can be transferred to a different site, especially if the map topology changes (cf. service robot in an office where the furniture has been moved). Pattern-based approaches tend to be used in non-safety critical applications, where explainability is less of an issue and where the environment is spatially constrained.

Planning-based approaches work well if goals, that the agents try to accomplish, can be explicitly defined and a map of the environment is available. In these cases, the planning-based approaches tend to generate better long-term predictions than the physics-based techniques and generalize to new environments better than the pattern-based approaches. 
In general, the runtime of planning-based approaches, based on classical planning algorithms ({i.e.} Dijkstra \citep{schrijver2012history}, Fast Marching Method \citep{sethian1996fast}, optimal sampling-based motion planners \citep{janson2018deterministic, karaman2011sampling}, value iteration \citep{littman1995complexity}) scales exponentially with the number of agents, the size of the environment
and the prediction horizon \citep{russell2016artificial}.
\subsubsection{On question 2:} 
In our view, \emph{Q2} is not confirmed. As we have seen, the different modeling approaches have various strengths and weaknesses. Although in principle it could be possible to incorporate the same contextual cues, there have been so far insufficient studies to compare prediction performance across modeling approaches. Moreover,
%
different modeling approaches exhibit varying degree of complexity and efficiency in including contextual cues from different categories.
Physics-based methods are by their very nature aware of the target agent cues and may be easily extended with other ones ({e.g.} social-force-based \citep{helbing1995social} and circular distribution-based \citep{coscia2018long}). 
Pattern-based methods can potentially handle all kind of contextual information which is encoded in the collected datasets. Some of them are intrinsically map-aware \citep{kucner2013conditional,bennewitz2005learning,roth2016iv}. Several others can be extended to include further types of contextual information ({e.g.} \cite{alahi2016social,trautman2010unfreezing,vemula2017socialattention,pfeiffer2018data,bartoli2017context}) but such extension may lead to involved learning, data efficiency and generalization issues ({e.g.} for the clustering methods \citep{bennewitz2005learning, chen2008pedestrian}). 
Planning-based approaches are intrinsically map- and obstacle-aware, natural to extend with semantic cues \citep{kitani2012activity,ziebart2009planning,Rudenko2018iros, Rhinehart_2018_ECCV}. Usually they encode the contextual complexity into an objective/reward function, which may fail to properly incorporate dynamic cues ({e.g.} changing traffic lights). Therefore, authors have to design specific modifications to include dynamic cues into the prediction algorithm (such as Jump Markov Processes in \citep{karasev2016intent}, local adaptations of the predicted trajectory in \citep{Rudenko2018iros, Rudenko2018icra}, game-theoretic methods in \citep{ma2016forecasting}. Unlike for the pattern-based approaches, target agents cues are natural to incorporate, e.g. as in \citep{kuderer2012feature, Rudenko2018icra, ma2016forecasting}, as both forward and inverse planning approaches rely on a dynamical model of the agents. Contextual cues-dependent parameters of the planning-based methods ({e.g.} reward functions for inverse planning and models for forward planning) are trivial and typically easier to learn but inference-wise less efficient for high-dimensional (target) agent states compared to the simple physics-based models.

\subsection{Application Domains}
\label{sec:application_areas_discussion}

In Sec.~\ref{sec:modeling_approaches_discussion} we have shown that all modeling approaches theoretically can handle various contextual cues. However, the question of preferring one approach over the others also depends on the task at hand.

\subsubsection{Service robots}
\label{sec:discussion:robotics}
Predictors for mobile robots usually estimate the most likely future trajectory of each person in the vicinity of the robot. The usual setup includes cameras, range and depth sensors mounted on the robot, operating on a limited-performance mobile CPU.

Physics-based or pattern-based human interaction models, capable of providing short-term high-confidence predictions (i.e. for 1-2 seconds), are best suited for local motion planning and collision avoidance in the crowd. Methods used to this end should have fast and efficient inference for predicting short-term dynamics of several people around the robot. In the simplest case, even linear velocity projection is sufficient for smoothing the robot's local planning \citep{Bai2015,chen2017decentralized}. More advanced methods should handle human-human interaction \citep{pellegrini2009you,ferrer2014behavior,alahi2016social,moussaid2010walking,Gupta2018SocialGAN}, the influence of robot's presence and actions on human motion \citep{oli2013human,schmerling2017multimodal,eiffert2019predicting,Rhinehart_2019_ICCV} and high-level body cues of human motion for disambiguating the immediate intention \citep{quintero2014pedestrian,unhelkar2015human,kooij2018ijcv,hasan2018mx}. In safety-critical applications, reachability-based methods provide a guarantee on local collision avoidance \citep{bansal2019hamilton}.  Furthermore, understanding local motion patterns is useful for compliant and unobstructive navigation \citep{palmieri2017kinodynamic,vintr2019time}.

For global path and task planning, on the other hand, long-term multi-hypothesis predictions ({i.e.} for 15-20 seconds ahead) are desired, posing a considerably more challenging task for the prediction system. Reactivity requirement is relaxed, however understanding dynamic \citep{ma2016forecasting,bera2017aggressive} and static contextual cues \citep{sun20173dof,kitani2012activity,chung2010mobile,coscia2018long}, which influence motion in the long-term perspective, reasoning on the map of the environment \citep{karasev2016intent,Rudenko2018icra} and inferring intentions of observed agents \citep{vasquez2016novel,best2015bayesian,rehder2017pedestrian} becomes more important.
For both local and global path planning, location-independent methods are best suited for predicting motion in a large variety of environments \citep{fernando2019neighbourhood,bansal2019hamilton,shi2019pedestrian}.

In terms of accuracy of the current state-of-the-art methods, experimental evaluations on simpler datasets, such as the ETH and UCY, show an average displacement error of \SI{0.19}{} -- \SI{0.4}{\metre} for \SI{4.8}{\second} prediction horizon \citep{yamaguchiCVPR2011,alahi2016social,vemula2017socialattention,radwan2018multimodal}. Linear velocity projection in these scenarios is estimated at \SI{0.53}{\metre} ADE. In more challenging scenarios of the ATC dataset with obstacles and longer trajectories an average error of \SI{1.4}{} -- \SI{2}{\metre} for \SI{9}{\second} prediction {has been reported} \citep{sun20173dof,alahi2016social,Rudenko2018iros}.

\subsubsection{Self-driving vehicles}
\label{sec:discussion:self-driving}
The early recognition of maneuvers of road users in canonical traffic scenarios is the subject of much interest in the self-driving vehicles application. Several approaches stop short of motion trajectory prediction ({i.e.} regression) and consider the problem as action classification, while operating on short image sequences. Sensors are typically on-board the vehicle, although some work involves infrastructure-based sensing ({e.g.} stationary cameras or laser scanners) which can potentially avoid occlusions and provide more precise object localization.

Most works consider the scenario of the laterally crossing pedestrian, dealing with the question what the latter will do at the curbside: start walking, continue walking, or stop walking \citep{schneider2013gcpr, keller2014tits, kooij2014eccv, kooij2018ijcv}. 
Some works enlarge the pedestrian crossing scenario, by allowing some initial pedestrian movement along the boardwalk before crossing (\cite{schneider2013gcpr} perform trajectory prediction, while other approaches are limited to crossing intention recognition, e.g. \citep{schneemann2016context,kohler2015stereo,fang2017board}). This scenario is safety-critical and crucial for autonomous vehicles to solve with high confidence. Pose and high-level contextual cues of the target agent \citep{kooij2018ijcv}, and the scene context modeling (e.g. location and type of the obstacles \citep{muench2019composable,volz2016predicting}, state of the traffic lights \citep{karasev2016intent}) are helpful to improve the crossing trajectory prediction.

As to cyclists, \cite{kooij2018ijcv} consider the scenario of a cyclist moving in the same direction as the ego-vehicle, and possibly bending left into the path of the approaching vehicle. \cite{pool2017iv} consider the scenario of a cyclist nearing an intersection with up to five different subsequent road directions. Both involve trajectory prediction.

For predicting motion of both cyclists and vehicles is it important to consider multi-modality and uncertainty of the future motion. Recently many authors have proposed solutions to this end \citep{chai2019multipath,zhao2019multi,hong2019rules,cui2019multimodal}. Furthermore, it is important to consider coordination of actions between the vehicles \citep{schmerling2017multimodal,Rhinehart_2019_ICCV}.

It is difficult to compare the experimental results, as the datasets are varying (different timings of same scenario, different sensors, different metrics). Several works report improvements vs. their baselines. For example, Fig. 2 in \citep{kooij2014eccv} shows that during pedestrian stopping, \SI{0.9}{} and \SI{1.1}{\metre} improvements in lateral position prediction can be reached with a context-based SLDS, compared to a simpler context-free SLDS and basic LDS (Kalman Filter), respectively, for prediction horizons up to \SI{1}{\second}. A live vehicle demo of this system at the ECCV'14 conference in Zurich, showed that the superior prediction of the context-based SLDS could lead to evasive vehicle action being triggered up to \SI{1}{\second} earlier, than with the basic LDS.

\subsubsection{{Surveillance}}
\label{sec:discussion:surveillance}
{The classification of goals and behaviors as well as the accurate prediction of human motion is of great importance for surveillance applications such as retail analytics or crowd control. Common setups for these applications use stationary sensors to monitor the environment.
While single-frame based systems allow to partially solve some tasks such as perimeter protection, incorporating a sequence of observations and making use of behavior prediction models often improve accuracy in cases of occlusions or measurements with low quality (e.g. noise, bad lighting conditions).}

{Traffic monitoring and management applications can benefit from from long-term prediction models, as they allow to associate
new observations with existing tracks ({e.g.} \cite{pellegrini2009you,yamaguchiCVPR2011,Luber2010,pellegrini2010improving}) and to model long-term distributions over possible future positions of each person 
\citep{yen2008goal,chung2012incremental}. 
Furthermore, it enables the analysis and control of customer flow in populated areas such as malls and airports, by gathering extensive information on human motion patterns \citep{ellis2009modelling,yoo2016visual,kim2011gaussian,tay2008modelling}, understanding crowd movement in light and dense scenarios, tracking individuals within them, and making future predictions of individuals or crowds (e.g. crowd density prediction). Often these methods benefit from employing sociological methods, such as understanding of social interaction, behavior analysis, group and crowd mobility modeling \citep{antonini2006behavioral,zhou2015learning,bera2016glmp,ma2016forecasting}}. 

Furthermore {identifying deviation from usual patterns often makes the foundation for anomaly detection methods that go beyond perimeter protection, as they analyze trajectories instead of the pure existence of a pedestrian in a specific region.}

Also in this application area it is difficult to compare results obtained by different approaches, due to the diversity of the used datasets and the way the evaluation has been performed (e.g. different prediction horizons). In terms of prediction accuracy, we report the most interesting results obtained in densely crowded environments using mainly image data. In these settings, recent state-of-the-art approaches achieve an average displacement error of \SI{0.08}{} -- \SI{1.2}{\metre} on the ETH, UC, NY Grand Central, Town Center and TrajNet datasets, and a final displacement error of \SI{0.081}{} -- \SI{2.44}{\metre}, with a prediction horizon that generally goes from \SI{0.8}{\second} up to \SI{4.8}{\second} (\cite{xue2018ss, xue2017bi, xue2019location, zhou2015learning, shi2019pedestrian}, the latter using a proprietary dataset and going up to a prediction horizon of \SI{10}{\second}).

\subsubsection{On question 3:} As we show in Sec.~\ref{sec:discussion:robotics}--\ref{sec:discussion:surveillance}, requirements to the motion prediction framework strongly depend on the application domain and particular use-case scenarios therein (e.g. vehicle merging vs. pedestrian crossing within the Intelligent Vehicles domain). Therefore, it is not possible to conclude achievement of absolute requirements of any sort. When considering concrete use-cases, industry-driven domains, such as intelligent vehicles (IV), appear to be the most mature in terms of formulated requirements and proposed solutions. For instance, requirements to the prediction horizon and metric accuracy for emergency braking of IV in urban driving scenarios are described in the \cite{ISO15622} standard, which defines norms for comfortable acceleration/deceleration rates for vehicles, conditioned on the maximum speed and traffic rules, as well as the distribution of pedestrian speed and acceleration. Therefore we conclude, that for specific use-cases, in particular for basic emergency braking for IV, solutions have achieved a level of performance that allows for industrialization into consumer products. Those use-cases can be considered solved. For other use-cases we expect more standardization and explicit formulation of requirements to take place in the near future. For instance, the standard for safety requirements for personal care robots \cite{ISO13482} suggests using sensors for detecting a human in the vicinity of the robot to issue a protective stop, and controlling the speed and force when the robot is in close proximity to humans to reduce the risk of collision. This standard, however, does not propose motion anticipation to improve the risk assessment.

Furthermore, several aspects of performance, robustness and generalization to new environments, discussed in the following sections, need to be explored before reaching further conclusions on maturity of the solutions. Finally, in order to reliably assess the quality of existing solutions across all application domains, is it critical to address the issues of benchmarking.

\subsection{Future Directions}
\label{sec:discussion:open_challenges}

Developing more sophisticated methods for motion prediction which go beyond Kalman filtering with simple motion models 
is a clear trend of the recent years. Modern techniques make extensive use of machine learning in order to better estimate context-dependent patterns in real-data, handle more complex environment models and types of motion, or even propose end-to-end reasoning on future motion from visual input. An increasing number of methods also includes reasoning on the global structure of the environment, intentions and actions of the agent. {Having these trends in mind, we see several directions of future research:}

\subsubsection{Use of {enhanced} contextual cues}

To analyze and predict human motion, as well as to plan and navigate alongside them, intelligent systems should have an in-depth semantic scene understanding. 
Context understanding with respect to features of the static environment and its semantics for better trajectory prediction is still a relatively unexplored area, see Sec.~\ref{sec:other_classifications:environment_description} for more details.



The same argument applies for the contextual cues of the dynamic environment. Socially-aware methods are making an important improvement over socially-unaware ones in such spaces where the target agent is not acting in isolation. However, most existing socially-aware methods still assume that all observed people are behaving similarly and that their motion can be predicted by the same model and with the same features.
Capturing and reasoning on the high-level social attributes is at an early stage of development, see Sec.~\ref{sec:other_classifications:target_agent} and Sec.~\ref{sec:other_classifications:interaction}, however recent methods take steps. Furthermore, most available approaches assume cooperative behavior, while real humans might rather optimize personal goals instead of joint strategies. In such cases, game-theoretic approaches are possibly better suited for modeling human behavior. Consequently, adopting classical AI and game-theoretic approaches in multi-agent systems is a promising research direction, that is only partly addressed in recent work, see e.g. \citep{ma2016forecasting, bahram2016game}.




One task where contextual cues become particularly important is long-term prediction of motion trajectories. While context-agnostic motion and behavioral patterns are helpful for short prediction horizons, long term predictions should account for intentions, based on the context and the surrounding environment.
Many pattern-based methods treat agents as particles, placed in the field of learned transitions, dictating the direction of future motion. Extending these models by more goal- or intention-driven predictions, that resemble human goal-directed behavior, would be beneficial for long-term predictions.

Consequently, further research on automatic goal inference based on the semantics of the environment is important. Most planning-based methods rely on a given set of goals, which makes them unusable or imprecise in a situation where no goals are known beforehand, or the number of possible goals is too high. Alternatively, one could consider identifying on-the-fly possible goals in the environment and predicting the way the agent may reach those goals. This would allow application of the planning-based methods in unknown environments. Additionally, semantic indicators of possible goals, coming from understanding the person's social role or current activity \citep{bruckschen2019human}, could lead to more robust intention recognition.

Apart from the contextual cues, discussed in this survey, there are many other factors influencing pedestrian motion, according to the recent studies \citep{rasouli2019autonomous}, e.g. weather conditions, time of day, social roles of agents. Future methods could benefit from closer connection to the studies of human motion and behavior in social spaces \citep{arechavaleta2008optimality,do2016group,gorrini2016age}.

\subsubsection{{Robustness and integration}}

Several practical aspects of deploying prediction systems in real environments should be considered in the future work.

Most of the presented methods are designed for specific tasks, scenarios or types of motion. These methods work well in certain situations, {e.g.} when prominent motion patterns exist in the environment, or when {the spatial structure of the environment and target agent's} goals are known beforehand. 
A conceptually interesting approach that uses a combination of multiple prediction algorithms to reason about best performance in the given situation is presented by \cite{lasota2017multiple}. The multiple-predictor framework opens a possibility for achieving more robust predictions when operating in undefined, changing situations, where a combination of strengths of different methods is required.

We suggest that more emphasis should be put on transfer learning and generalization of approaches to new environments. 
Learning and reasoning on basic, invariant rules and norms of human motion and collision avoidance is a better approach in this case. When having access to several environments, domain adaptation could be potentially used for learning generalizable models.

Integration of prediction in planning and control is another worthwhile {topic for overall system robustness}. Predicting human motion is usually motivated with increased safety of human-robot interaction and efficiency of operation. However, the insights on exploiting predictions in the robot's motion or action planning module are typically left out of scope in many papers. Future work would benefit from outlining possible ways to incorporate predictions in the robot control framework.

\section{Conclusions}
\label{sec:conclusions}

In this work we present a thorough analysis of the human motion trajectory prediction problem. We survey the literature across multiple domains and propose a taxonomy of motion prediction techniques. Our taxonomy builds on the two fundamental aspects of the motion prediction problem: the model of motion and the input contextual cues. We review the relevant trajectory prediction tasks in several application areas, such as service robotics, self-driving vehicles and advance surveillance systems. Finally, we summarize and discuss the state of the art along the lines of three major questions and outlined several prospective directions of future research.


\emph{``Prediction is very difficult, especially about the future''}. This quote (whose origin has been attributed to multiple people) certainly remains applicable to motion trajectory prediction, despite two decades of research and the 200+ prediction methods listed in this survey. We hope that our survey increases visibility in this rapidly expanding field and the will stimulate further research along the directions discussed.


    




\begin{acks}
Authors would like to thank Achim J. Lilienthal for valuable feedback and suggestions.
\end{acks}

\begin{funding}
This work has been partly funded from the European Union's Horizon 2020 research and innovation programme under grant agreement No 732737 (ILIAD).
\end{funding}





\bibliographystyle{SageH}
\bibliography{main-sage.bib}

\end{document}